\documentclass[acmsmall]{acmart}

\usepackage{booktabs} 
\usepackage{hyperref}   
\usepackage{url}  
\usepackage{multirow}  
\usepackage{color} 
\usepackage{tabularx}
\usepackage{makecell}
\usepackage{appendix}
\usepackage{lineno}
\usepackage{diagbox}
\usepackage{amsmath}
\usepackage{arydshln} 

\usepackage{algorithmic}

\usepackage[linesnumbered,ruled]{algorithm2e} 

\SetAlFnt{\small}
\SetAlCapFnt{\small}
\SetAlCapNameFnt{\small}
\SetAlCapHSkip{0pt}
\IncMargin{-\parindent}
\usepackage{longtable}

\usepackage{soul}
\usepackage{color, xcolor} 
\sethlcolor{yellow}
\soulregister{\cite}7 
\soulregister{\citep}7 
\soulregister{\citet}7 
\soulregister{\ref}7 
\soulregister{\pageref}7 
\soulregister{\underline}7 

\acmJournal{TALLIP}

\setcopyright{acmlicensed}

\begin{document}

\title{ADKGD: Anomaly Detection in Knowledge Graphs with Dual-Channel Training}

\author{Jiayang Wu}
\affiliation{ 
	\institution{Jinan University}
	\city{Guangzhou}
	\country{China}
}
\email{csjywu1@gmail.com}

\author{Wensheng Gan}
\authornote{This is the corresponding author.}
\affiliation{
	\institution{Jinan University}
	\city{Guangzhou}
	\country{China}
}
\email{wsgan001@gmail.com}

\author{Jiahao Zhang}
\affiliation{ 
	\institution{South China Normal University}
	\city{Guangzhou}
	\country{China}
}
\email{jerrycu@foxmail.com}

\author{Philip S. Yu}
\affiliation{
	\institution{University of Illinois Chicago}
	\city{Chicago}
	\country{USA}
}
\email{psyu@uic.edu}

\begin{abstract}
  In the current development of large language models (LLMs), it is important to ensure the accuracy and reliability of the underlying data sources. LLMs are critical for various applications, but they often suffer from hallucinations and inaccuracies due to knowledge gaps in the training data. Knowledge graphs (KGs), as a powerful structural tool, could serve as a vital external information source to mitigate the aforementioned issues. By providing a structured and comprehensive understanding of real-world data, KGs enhance the performance and reliability of LLMs. However, it is common that errors exist in KGs while extracting triplets from unstructured data to construct KGs. This could lead to degraded performance in downstream tasks such as question-answering and recommender systems. Therefore, anomaly detection in KGs is essential to identify and correct these errors. This paper presents an anomaly detection algorithm in knowledge graphs with dual-channel learning (ADKGD). ADKGD leverages a dual-channel learning approach to enhance representation learning from both the entity-view and triplet-view perspectives. Furthermore, using a cross-layer approach, our framework integrates internal information aggregation and context information aggregation. We introduce a kullback-leibler (KL)-loss component to improve the accuracy of the scoring function between the dual channels. To evaluate ADKGD's performance, we conduct empirical studies on three real-world KGs: WN18RR, FB15K, and NELL-995. Experimental results demonstrate that ADKGD outperforms the state-of-the-art anomaly detection algorithms. The source code and datasets are publicly available at \url{https://github.com/csjywu1/ADKGD}.
\end{abstract}

%
%



\maketitle

\renewcommand{\shortauthors}{J. WU \textit{et al.}}

\section{Introduction}  \label{sec: introduction}

Knowledge graphs (KGs) are a data structure that could effectively integrate numerous real-world relations in triplets \cite{sheth2019knowledge}. They integrate data from diverse sources into a unified structure. Knowledge graphs are advanced data structures that represent entities and their interrelations in a graph format, where nodes denote entities and edges represent relationships. This structured representation allows for effective storage and retrieval. Based on the deeper semantic understanding of the data, they can be extensively performed in many downstream tasks, such as question answering \cite{huang2019knowledge, saxena2020improving} and recommender systems \cite{li2024ripple}. Question-answering systems could provide a structured way to interpret and answer complex queries. For instance, a system can use a KG to answer a question like ``What are the main contributions of Albert Einstein?" by linking directly to relevant data points such as his theories, publications, and awards, thus offering precise and comprehensive answers. In the field of recommendation systems, KG can help identify buying patterns and customer preferences by linking products with customer demographics and purchasing history. Then, this could allow for more effective strategies in marketing and inventory management.

However, extracting triplets from real-world unstructured data in practice could introduce numerous errors. These errors could lead the downstream task to perform worse. Therefore, anomaly detection should be necessarily considered after constructing KGs. Anomaly detection in knowledge graphs is a task to identify unusual or unexpected patterns in the data that deviate from the norm, which may indicate errors \cite{jia2018pattern}. It is closely related to other tasks, such as KG completion and refinement \cite{lin2015learning}. While completion focuses on filling in gaps and refinement on improving quality, anomaly detection serves as a quality assurance mechanism, ensuring that additions and modifications made through completion and refinement do not introduce errors. In essence, anomaly detection acts as a safeguard, maintaining the integrity and reliability of the knowledge graphs.

\begin{figure}[ht]
    \centering
    \includegraphics[clip,scale=0.3]{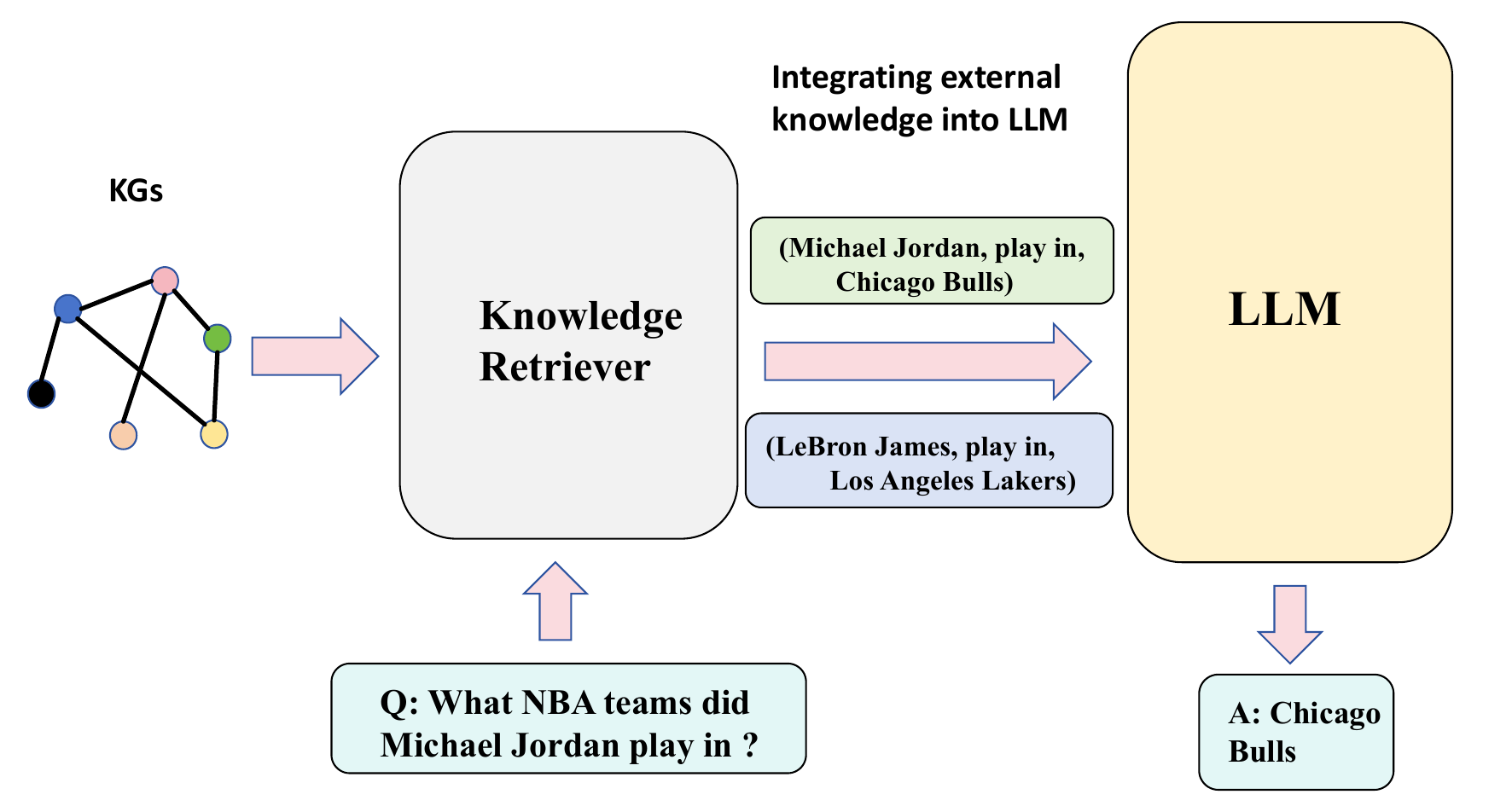}
    \caption{An example of utilizing KGs to retrieve external knowledge to enhance the LLMs generation.}
    \label{fig:llmkg}
\end{figure}

In recent years, large language models (LLMs) have seen significant development and application \cite{wu2023ai, wu2023multimodal, ouyang2022training}. However, these models often suffer from the hallucination phenomenon \cite{huang2023survey, xu2024hallucination}, where they generate text that sounds plausible but is inaccurate or irrelevant. This issue arises mainly due to the knowledge gaps in the training data. Using KGs as an external information source can effectively mitigate this problem \cite{fei2021enriching}.  The combination with KGs can effectively reduce hallucinations in LLMs, so the accuracy of the information within the KGs is crucial. Knowledge graph anomaly detection can identify and correct errors within the graph, ensuring that the knowledge base used by LLMs is reliable. If the knowledge graph contains erroneous information, LLMs might generate misleading content, affecting user experience and trust. The accuracy of knowledge graphs not only improves the safety and consistency of the content generated by LLMs but also enhances their reasoning capabilities \cite{lewis2020retrieval, lin2019kagnet, li2023graph, wang2023boosting}. For example, as shown in Fig. \ref{fig:llmkg}, when asked "Did Michael Jordan play for the Los Angeles Lakers?", LLM might generate an incorrect answer that contradicts known facts. The related knowledge retrieved using the KGs is fed into LLM as additional contextual information, enabling LLM to use the retrieved knowledge to generate more accurate responses. Additionally, for complex questions that involve multi-hop relationships and attribute comparisons, precise reasoning is essential. KoPL programming language translates natural language into basic function combinations for complex problem-solving \cite{yao2023viskop}. For instance, to determine who is taller between LeBron James and his father, the process involves querying their heights and comparing them. This method showcases rigorous reasoning, benefiting human-machine interaction and improving response interpretability \cite{petroni2019language}.

Due to the complex interactions in real-world data, detecting anomalies in knowledge graphs becomes challenging. The methods in KG anomaly detection can be classified into three primary forms: rule-based, path-based, and embedding-based methods. The rule-based methods detect errors by relying on predefined rules \cite{guo2018knowledge, pellissier2017completeness}. KG errors are defined as statements that violate any of these rules. However, these methods are not generalizable because different KGs require distinct sets of rules based on domain-specific knowledge. This dependence on specific rules restricts their applicability across different KGs, reducing their versatility in addressing the diverse range of errors encountered in real-world scenarios. Path-based methods focus on the paths between entities within the KG. These methods analyze the connections and relationships along different paths to identify anomalies. While path-based methods offer a more dynamic approach compared to rule-based methods, they still face challenges in distinguishing between semantically correct and incorrect paths. Embedding-based methods, which have gained significant attention recently, are based on the representations of entities and relations within the KG \cite{jia2019triple,melo2017detection}. These methods use embeddings to detect anomalies by analyzing the data's underlying patterns. Embedding-based approaches are more flexible and can generalize better across different KGs compared to rule-based methods. They often employ naive negative sampling to create synthetic labels for unsupervised anomaly detection. However, these methods also face challenges due to the diverse and often unlabeled nature of real-world KG errors.

Zhang \textit{et al.} \cite{zhang2022contrastive} proposed a model called CAGED that could consider internal and context information of the graph. Aggregating internal information is often referred to as the base layer. Aggregating context information corresponds to the additional layer. The cross-layer training process effectively integrates these two layers, ensuring that both the detailed internal structure and the broader contextual information are considered simultaneously. This integration enhances the detection performance by leveraging the strengths of both the internal and external perspectives of the graph data. Therefore, we continue to use the cross-layer training process and further utilize dual-channel training in our framework. It includes entity-view learning in channel I and triplet-view learning in channel II. Moreover, to ensure the dual channel is effective, we introduce the kullback-leibler (KL)-loss into our framework to ensure better learning from the scoring function from both channels. Our contributions can be summarized as follows:

\begin{itemize}
    \item  We developed a dual-channel learning framework, namely ADKGD, which could learn the representations with entity-view and triplet-view.
	
    \item  We perform effective anomaly detection in our framework by integrating internal and context information aggregation through cross-layer learning.
	
    \item  To improve the accuracy of the scoring function, ADKGD introduces a KL-loss component between the dual channels.
	
    \item We conduct empirical studies on three real-world KGs. Experimental results demonstrate that ADKGD outperforms state-of-the-art anomaly detection algorithms.
\end{itemize}

The remaining parts of this paper are given as follows. Related work is stated and summarized in Section \ref{sec: relatedwork}. The preliminaries and basic knowledge are described in Section \ref{sec: preliminaries}. Our method ADKGD is detailed in Section \ref{sec: algorithm}. Furthermore, experimental results are shown in Section \ref{sec: experiments}, and the conclusion is presented in Section \ref{sec: conclusion}.

\section{Related Work} \label{sec: relatedwork}
\subsection{Triplets scoring function}

The triplets scoring function can measure the compatibility between entities and their relationships. Several scoring functions have been developed to assess the score of triplets in KGs. Common methods include distance-based scoring functions and semantic matching scoring functions. A prominent example of a distance-based scoring function is TransE \cite{bordes2013translating}. TransE minimizes the distance between the head entity, relation, and tail entity embeddings: $\| \mathbf{h} + \mathbf{r} - \mathbf{t} \|$, where $ \mathbf{h}, \mathbf{r}, \mathbf{t} $ are the embeddings of the head entity, relation, and tail entity, respectively. DistMult \cite{yang2015embedding} uses a bilinear scoring function, which is given by: $f_r(h, t)$ = $\langle \mathbf{h}, \mathbf{r}, \mathbf{t} \rangle$, where $ \langle \mathbf{h}, \mathbf{r}, \mathbf{t} \rangle $ denotes the trilinear dot product. ComplEx \cite{trouillon2016complex} extends the idea of DistMult to complex numbers. Its scoring function is: $f_r(h, t)$ = $\textit{Re}(\langle \mathbf{h}, \mathbf{r}, \overline{\mathbf{t}} \rangle)$, where $ \textit{Re} $ denotes the real part and $ \overline{\mathbf{t}} $ denotes the complex conjugate of the tail entity vector. Negative samples are crucial for training these models. A common approach is to generate negative samples by randomly replacing the head or tail entities in positive triplets. For instance, given a positive triplet $(h, r, t)$, a negative triplet could be $(h', r, t)$ or $(h, r, t')$, where $h'$ or $t'$ are randomly selected entities. Therefore, scoring functions can also be used for anomaly detection in KGs. By calculating the score of each triplet, we can identify anomaly triplets with low scores, which often represent unlikely relationships in KGs.

\subsection{Anomaly detection methods in KGs}

The traditional KG embedding scoring functions such as TransE \cite{bordes2013translating}, DistMult \cite{yang2015embedding}, and ComplEx \cite{trouillon2016complex} do not consider the errors in KGs and thus cannot learn discriminative representations for anomaly triplets. Therefore, several advanced methods have been proposed to detect anomalies in KGs, including CKRL \cite{shan2018confidence}, KGTtm \cite{jia2019triple}, KGIst \cite{belth2020normal}, and CAGED \cite{zhang2022contrastive}. These methods employ unsupervised learning techniques to improve the detection of anomaly triplets. For simultaneous noise detection, CKRL defines local and global confidence scores based on the internal structure of KGs. It optimizes global consistency with confidence-weighted translations, enhancing KG embedding robustness \cite{shan2018confidence}. Building on confidence and trustworthiness, KGTtm quantifies the semantic correctness and factual accuracy of triplets. It uses a cross-neural network to measure trustworthiness at entity, relation, and KG levels, allowing comprehensive evaluation \cite{jia2019triple}. KGIst uses unsupervised inductive summarization to represent KGs, focusing on learning "normal" patterns. It employs the minimum description length principle to identify anomalies \cite{belth2020normal}. The CAGED model adds contrastive learning to KG modeling and checks the triplet credibility by using both KG embedding and contrastive learning loss. It leverages multi-view and internal triplet consistency \cite{zhang2022contrastive}.

Cross-layer learning has demonstrated significant advantages in various algorithms. For example, LGLP \cite{cai2021line} transforms the original graph into a line graph, allowing the link prediction task to be directly conducted within the line graph. This approach avoids information loss in graph pooling operations and performs well in both sparse and dense graphs. In addition, LGCL \cite{zhang2023line} enhances model robustness by maximizing the similarity between subgraphs and line graphs through contrastive learning. This cross-layer learning approach uses different levels of graph representations to enhance feature learning, enabling the model to better handle sparse and complex network data. The previous methods are used for link prediction, while the CAGED method utilizes cross-layer learning for anomaly detection \cite{zhang2022contrastive}. CAGED also integrates graph representations from different views to identify anomalies in KGs.

\subsection{Dual-channel training}


Dual-channel training arises from the desire to improve item transition modeling, both within and across sessions \cite{fei2022matching}. This approach enhances the recommendation systems' ability to capture collaborative information and similar behavior patterns. It is particularly beneficial in the context of session-based recommendations, where the data is often anonymized and only the user's actions within a session are available. Dual-channel training uses two types of information: the intra-session channel, which captures transitions within the current session, and the inter-session channel, which captures transitions from neighboring sessions. This dual approach allows for a more comprehensive understanding of user behavior by integrating signals from both the target session and similar past sessions.

Dual-channel training has been widely used in graph learning algorithms, particularly in the development of graph neural networks (GNNs). DGTN is an innovative method designed to model item transitions across different sessions to enhance the performance of session-based recommendation systems \cite{zheng2020dgtn}. Traditional methods only consider item transitions within the target session, ignoring the complex transitions in neighboring sessions. DGTN integrates the target session and its similar neighboring sessions into a single graph and explicitly injects transition signals into the embeddings through channel-aware propagation. D2PT is designed to address the issue of graph learning with weak information (GLWI) \cite{liu2023learning}. D2PT introduces a dual-channel architecture that alleviates isolated node problems through an enhanced global graph, further improving information propagation. DualGCL \cite{luo2023dual} is a graph contrastive learning method that enhances the performance of graph embeddings through a dual-channel structure. This method achieves superior classification accuracy across multiple benchmark datasets compared to existing contrastive learning models, demonstrating its advantages in capturing graph structure information and node representations.

\section{Preliminaries}  \label{sec: preliminaries}

In this section,  we first provide the formal problem statement for detecting anomalies in a knowledge graph. Then, the definitions and fundamental concepts used in this paper are introduced.

\begin{definition}[Problem definition]
    \rm A knowledge graph can be formulated as $G$ = $(E, R, T)$, where $E$ denotes the set of entities, $R$ represents the set of relations between these entities,  and $T$ comprises all the triplets within the KG, each triplet being a statement of the form $(e_{1}$, $r$, $e_{2})$ indicating a relationship $r \in R$ between $e_1$, $e_2 \in E$. The whole KG contains the label triplets as $Y$ = $\{(t_i, y_i)\}^N_{i=1}$ where $t_i$ $\in $ $T$. It consists of two types of data, where $y_i \in \{0, 1\}$ indicates the correctness of the triplet. The total training process is under unsupervised training, and the labels are inaccessible during training. During the training process, we treat all the samples as positive and use them to construct the negative samples. In our model, we use these samples to learn the representations and how to score the anomalies, i.e., establish the scoring function. In the testing step, it directly uses the representations and scoring function to determine the degree of the node's anomalies. This framework, represented as  as $\mathcal{F} \left( f\left( G;\varTheta \right) , Y;\varPhi \right)$, learns the representations from $f(\cdot ; \varTheta)$ and constructs the anomaly scoring function from $\mathcal{F}(\cdot ; \varPhi)$ with  learnable parameters $\varTheta$ and $\varPhi$. By optimizing the loss over the training data, we aim to learn the scoring function that most effectively identifies anomaly triplets in $G$. Then, it uses a ranking method to predict the label $\hat{Y}$  for $G$ and compares these predictions against the actual ground truth labels  $Y$. 
\end{definition}

\begin{definition}[\textit{BI-LSTM}]
    \rm  Long short-term memory (LSTM) networks have become essential tools for handling sequential data tasks, including natural language processing and time series analysis, due to their ability to manage long-term dependencies. A notable enhancement of LSTM networks is the bidirectional LSTM (\textit{BI-LSTM}), which enhances context comprehension by processing data sequences in both forward and backward directions. By employing two separate hidden states for these directions, \textit{BI-LSTM} networks effectively capture contextual information from both past and future states within the data \cite{zhou2016attention}. In our model, we utilize two types of \textit{BI-LSTM} networks: one with entity-view and the other with triplet-view. The \textit{BI-LSTM} with entity-view maintains the same dimensionality for the input and output sequences, preserving the structure of the data across the layers. Formally, we define the process as follows:
    \begin{equation} \label{eq1}
       \tilde{e}_h, \tilde{e}_r, \tilde{e}_t = \textit{BI-LSTM}(e_h, e_r, e_t), 
        q_i = [\tilde{e}_h; \tilde{e}_r; \tilde{e}_t],
    \end{equation}
    where $ e_h, e_r, e_t $ are the input embeddings of the head entity, relation, and tail entity, respectively, and $ \tilde{e}_h, \tilde{e}_r, \tilde{e}_t $ are their corresponding output embeddings. On the other hand, the \textit{BI-LSTM} with triplet-view reduces the dimensionality of the input sequences, compressing the information into lower-dimensional representations. The process is defined as follows:
    \begin{equation} \label{eq2}
        \tilde{e}_{h,r,t} = \textit{BI-LSTM-D}(e_h, e_r, e_t), 
            q_i' = [\tilde{e}_{h,r,t}],
    \end{equation}
    where $ e_h, e_r, e_t $ are the input embeddings of the head entity, relation, and tail entity, respectively, and $\tilde{e}_{h,r,t}$ are their corresponding output embeddings.       
\end{definition}

\begin{definition}[Graph encoder layer]
    \rm Unlike \textit{BI-LSTM}, which learns representations within the intra-view of the triplets, the graph encoder layer focuses on aggregating information from the head neighbor triplet or tail neighbor triplet for each anchor triplet \cite{wu2020comprehensive}. Given an anchor triplet $q_i$, we update its embedding representation based on the weighted aggregation of its neighbor triplets, e.g., \{$q_1$, $q_2$, $q_3$, $\cdots$, $q_j$\}. The neighbor triplets can be divided into two types: head neighbor triplet or tail neighbor triplet. A head neighbor triplet shares the same head entity as the anchor triplet. For example, if the anchor triplet is $(h, r, t)$, then a head neighbor triplet would be $(h, r', t')$, where the head entity $ h $ remains the same while the relations and tail entities differ.
    Conversely, a tail neighbor triplet shares the same tail entity as the anchor triplet. For example, if the anchor triplet is $(h, r, t)$, then a tail neighbor triplet would be $(h', r', t)$, where the tail entity $ t $ remains the same while the head entities and relations differ.
    
    The weight between the anchor triplet $q_i$ and its neighbor triplet $q_j$ is calculated as follows:
    \begin{equation}
    \hat{\alpha}_{ij} = \textit{sim}(q_i, q_j).
    \end{equation}
    $\hat{\alpha}_{ij}$ indicates the importance of triplet $j$ to triplet 
$i$. Here $\textit{sim}$ is the attentional function: 
    $\mathbb{R}^n \times \mathbb{R}^n \rightarrow \mathbb{R}$. To make attention scores easily comparable across different triplets, we normalize them by applying a \textit{softmax} function:
    \begin{equation}
    \alpha_{ij} = \frac{\exp(\hat{\alpha}_{ij})}{\sum_{k=1}^{m} \exp(\hat{\alpha}_{ik})}.
    \end{equation}  
       where $m$ is the number of the neighbors. The head neighbor triplet aggregation can be calculated with a sigmoid function, as depicted:
    \begin{equation}
    {x}_i = \sigma \left( \sum_{j=1}^{m} \alpha_{ij} q_j \right),
    \end{equation}
\end{definition}

\begin{definition}[Knowledge graph scoring function]
     \rm The knowledge graph scoring function evaluates the score of a given triplet $(h, r, t)$, where $h$ is the head entity, $r$ is the relation, and $t$ is the tail entity. This function calculates a score for each triplet based on the embeddings of the head entity, relation, and tail entity. Higher scores indicate a higher compatibility of the triplet. Given the embeddings of the head entity $\mathbf{e}_h \in \mathbb{R}^d$, relation $\mathbf{e}_r \in \mathbb{R}^d$, and tail entity $\mathbf{e}_t \in \mathbb{R}^d$, the scoring function can be defined as follows:
    \begin{equation}
        f(h, r, t) = \phi(\mathbf{e}_h, \mathbf{e}_r, \mathbf{e}_t),
    \end{equation}
    where $\phi(\mathbf{e}_h, \mathbf{e}_r, \mathbf{e}_t)$ is a scoring function that measures the compatibility of the embeddings. One common choice for $\phi$ is the TransE scoring function \cite{bordes2013translating}:   
    \begin{equation}
        \phi(\mathbf{e}_h, \mathbf{e}_r, \mathbf{e}_t) = -\|\mathbf{e}_h + \mathbf{e}_r - \mathbf{e}_t\|_2,
    \end{equation}
    where $\|\cdot\|_2$ denotes the $L_2$ norm. In this case, the goal is to minimize the distance between $\mathbf{e}_h + \mathbf{e}_r$ and $\mathbf{e}_t$.
    To incorporate anomaly detection, we consider that the higher the value of the scoring function, the more likely the triplet is an anomaly. This can be used to identify potential errors or anomalies within the knowledge graph. For example, if $f(h, r, t)$ exceeds a certain threshold, the triplet $(h, r, t)$ can be flagged as a potential anomaly.
    The scoring functions are used to train the embeddings $\mathbf{e}_h$, $\mathbf{e}_r$, and $\mathbf{e}_t$ such that correct triplets $(h, r, t)$ receive lower scores compared to anomaly ones. Conversely, triplets with anomaly patterns will have higher values of $f(h, r, t)$, indicating they are less compatible.
\end{definition}

\begin{definition}[Consistency loss function]
    \rm The consistency loss function is introduced to ensure that the vector representations obtained from two different views are consistent with each other. This is particularly useful when learning from multiple perspectives, as it encourages the model to produce similar embeddings across these views. To achieve this, we can use the kullback-leibler (KL) divergence to measure the similarity between the representations. Given two matrix representations $Q_1$ and $Q_2$, the consistency loss $\mathcal{L}_{\textit{consistency}}$ is defined as follows \cite{hou2017deep}:

    \begin{equation}
        \mathcal{L}_{\textit{consistency}} = \textit{KL}(Q_1 || Q_2) = \sum_{i} Q_1^{(i)} \log \left( \frac{Q_1^{(i)}}{Q_2^{(i)}} \right),
    \end{equation}
    where $Q_1^{(i)}$ and $Q_2^{(i)}$ are the $i$-th elements of the matrix representations $Q_1$ and $Q_2$, respectively. The KL divergence $\textit{KL}(Q_1 || Q_2)$ quantifies the difference between the two probability distributions represented by $Q_1$ and $Q_2$. The purpose of the consistency loss is to minimize the divergence between these two matrix representations, thereby encouraging the model to produce consistent results with different views. This helps in maintaining coherence and reliability in the learned representations, making them more robust and interpretable.
\end{definition}

\section{Algorithm} \label{sec: algorithm}

\begin{figure}[ht]
    \centering
    \includegraphics[clip,scale=0.4]{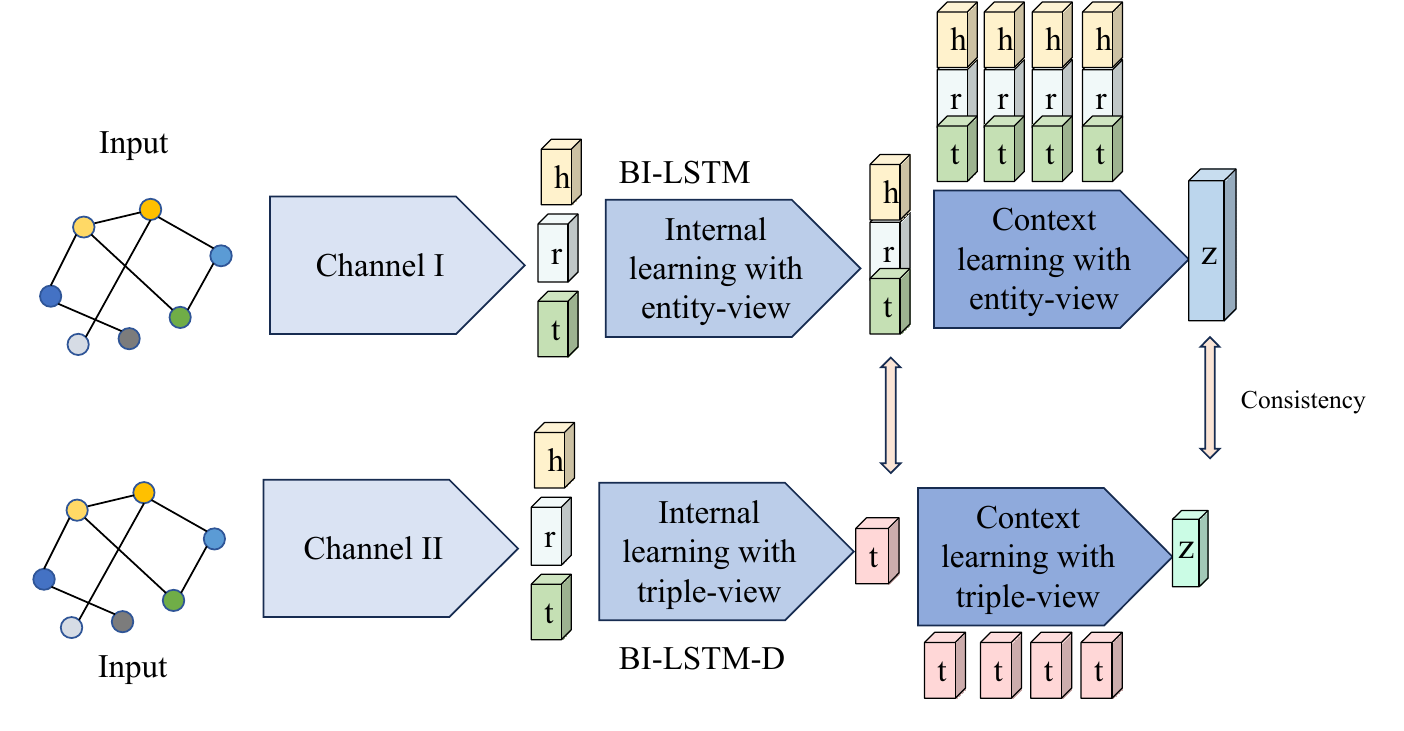}
    \caption{The framework of ADKGD. Channel I represents the entity-view, where internal learning is conducted using BI-LSTM. In this process, the input and output dimensions remain consistent, ensuring that the embeddings of entities and relations retain their original dimensionality. Channel II represents the triple-view, where internal learning is performed using BI-LSTM-D. Unlike Channel I, this process reduces the dimensionality of the triplet embeddings, allowing for a more compact representation. The complementary nature of these views, combined with their respective neighbor aggregation strategies, enhances the detection of anomalous patterns in the knowledge graph.    Both channels operate on distinct views of the knowledge graph, with Channel I focusing on entities and Channel II emphasizing triples. The outputs of these two views are aligned using a consistency loss, ensuring that both perspectives contribute to a unified and robust anomaly detection model. }
    \label{fig:framework}
\end{figure}

Our framework learns to detect anomalies in KG with dual-channel training. As shown in Figure \ref{fig:framework}, each channel of ADKGD is mainly composed of two parts. The first part is to learn the representations from the cross-layer. The cross-layer learning can decompose into internal information aggregation and context information aggregation. The internal information aggregation is to use \textit{BI-LSTM} to learn the internal relationship for each triplet in KG. However, there is a difference between channel I and channel II. Channel I uses the entity-view, while channel II uses the triplet-view. We use consistency loss to ensure the learning is effective between these two channels. The context information aggregation is to aggregate the representations with their neighbor triplets, including head neighbor and tail neighbor triplets. It's also different between the two channels. During the training, it uses methods the same as TransE to construct the negative triplets. It learns to calculate the score for positive samples and negative samples. Its goal is to make the score between the positive samples and negative samples with a margin. Finally, we could use these representations and score functions to identify the anomalies in the unlabeled data.

\subsection{Data preparation}

\begin{figure}[ht]
    \centering
    \includegraphics[clip,scale=0.7]{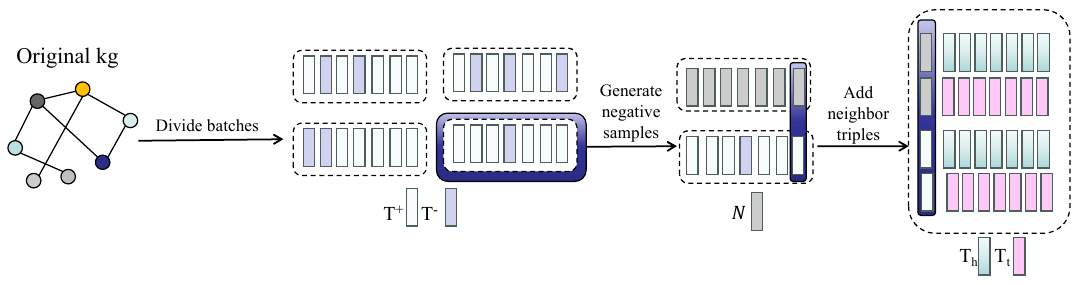}
    \caption{The data preparation for training. Starting from the original knowledge graph (KG), the data is divided into batches, $ B_1, B_2, \ldots, B_n $, each containing triplets $ T $. These triplets are split into positive ($ T^+ $) and negative ($ T^- $) samples. Negative samples ($ N $) are generated by replacing the head ($ h $) or tail ($ t $) entity of $ T^+ $ with a random entity from the KG, ensuring they do not exist in the original graph. For each triplet, neighbor triplets are added to provide contextual information. Head neighbor triplets ($ T_h $) share the same head entity, while tail neighbor triplets ($ T_t $) share the same tail entity. The final structure of each batch includes positive and negative triplets, as well as their neighbors ($ T_h $ and $ T_t $), enabling the model to learn both internal relationships and broader context for effective anomaly detection.}
    \label{fig:data}
\end{figure}

As shown in Figure \ref{fig:data}, we start with the original Knowledge Graph (KG) data, which is divided into multiple batches, denoted as $ B_1, B_2, \ldots, B_n $. This batching process helps manage the computational load and facilitates efficient training of the model. Each batch $ B_i $ contains a subset of the triplets from the original KG. These triplets $T$ include both correct triplets ($ T^+ $) and anomalies ($T^-$).

Within each batch, we generate negative samples ($N$) for each triplet ($T$). This ensures that both correct triplets and anomalies have their corresponding negative samples. For each triplet $T$ in the batch, we generate corresponding negative samples ($N$) by randomizing parts of the positive triplets. Specifically, we create negative samples by either replacing the head entity $h$ or the tail entity $t$ with a random entity from the KG. This randomization helps the model learn to identify and distinguish incorrect relationships. The process can be described as follows, where $G$ represents the set of triplets in the knowledge graph:
\[
N = \{ (h', r, t) \mid h' \neq h \textit{ and } (h', r, t) \notin G \} \cup \{ (h, r, t') \mid t' \neq t \textit{ and } (h, r, t') \notin G \}.
\]

For each triplet, we obtain head neighbor triplets ($ T_h $) and tail neighbor triplets ($ T_t $) from the graph. These neighbor triplets provide additional context for the model to learn from the relationships surrounding each entity in the triplet. The definitions are as follows:
\begin{itemize}
    \item \textbf{Head neighbor triplet} ($ T_h $): For each triplet $ (h, r, t) $, we identify and include triplets that share the same head entity $ h $. $T_h$ = $\{ (h, r', t') \mid (h, r', t') \in \textit{G} \}$.
    \item \textbf{Tail neighbor triplet} ($ T_t $): For each triplet $ (h, r, t) $, we identify and include triplets that share the same tail entity $ t $. $T_t$ = $\{ (h', r', t) \mid (h', r', t) \in \textit{G} \}$.
\end{itemize}

The final dataset structure for each batch includes the samples $ T $ and their corresponding negative samples $ N $, along with their head neighbor triplets $ T_h $ and tail neighbor triplets $ T_t $. This structured data is then fed into the model for training. This process enables the model to learn to distinguish between correct and anomalous triplets by leveraging both the triplets themselves and their neighboring contexts.

\subsection{Entity-view for detecting anomalies}

In this part, we utilize the entity-view in CAGED \cite{zhang2022contrastive} as one learning view for detecting anomalies. The process involves two main stages: internal information aggregation using \textit{BI-LSTM} and aggregation with neighbor triplets, as shown in Figure \ref{fig:hebing} (A). First, we use \textit{BI-LSTM} to capture the relationship of the entities and relations within a triplet. This approach helps in understanding the context by processing the sequences in both forward and backward directions. The dimension can be denoted as:
$
\mathbb{R}^n \times \mathbb{R}^n \times \mathbb{R}^n \rightarrow \mathbb{R}^n \times \mathbb{R}^n \times \mathbb{R}^n.
$
Formally, given the embeddings of the head entity $e_h$, relation $e_r$, and tail entity $e_t$, the \textit{BI-LSTM} outputs can be expressed as:
\begin{equation}
\tilde{e}_h, \tilde{e}_r, \tilde{e}_t = \textit{BI-LSTM}(e_h, e_r, e_t), \forall (h, r, t) \in G,
\end{equation}
where $G$ represents the set of triplets in the knowledge graph. We concatenate these outputs to form a single vector representation for each triplet:
\begin{equation}
q_i = [\tilde{e}_h; \tilde{e}_r; \tilde{e}_t].
\end{equation}

Next, we aggregate these representations with their neighbor triplets to derive more robust embeddings. There are two types of neighbor triplets considered: head neighbor triplets and tail neighbor triplets. Firstly, for each triplet, we compute the similarity with multiple head neighbor triplets. The similarities are then processed using the \textit{softmax} function to normalize. The aggregation is performed by weighing the neighbor triplets according to these normalized similarities. This process results in obtaining  $z_1$:
\begin{equation}
z_1 = \sum_{j} \textit{softmax}(\textit{sim}(q_i, \textit{q}_j)) \cdot \textit{
q}_j,
\end{equation}
where $\textit{sim}$$(q_i, q_j)$ denotes the similarity between the anchor triplet $q_i$ and a head neighbor triplet $q_j$. The similarity is calculated by using the dot product. The \textit{softmax} function is used to normalize these similarities, ensuring they sum to 1 across all neighbor triplets. This weighted aggregation results in the embedding $z_1$, which captures the context from the head neighbors. Similarly, for each triplet, we compute the similarity with multiple tail neighbor triplets. These similarities are also normalized using the \textit{softmax} function. The aggregation is done by weighing the neighbor triplets according to the normalized similarities. This process results in obtaining $z_2$:
\begin{equation}
z_2 = \sum_{k} \textit{softmax}(\textit{sim}(q_i, \textit{q}_k)) \cdot \textit{q}_k.
\end{equation}

\begin{algorithm}
    \small
    \caption{entity-view anomaly detection}
    \label{alg:entity_view}
    \begin{algorithmic}[1]
    \renewcommand{\algorithmicrequire}{\textbf{Input:}}
    \renewcommand{\algorithmicensure}{\textbf{Output:}}
    \REQUIRE{a knowledge graph $G$; the number of epochs $k$.}
    \ENSURE{node representations: $q$, $z_1$, and $z_2$.} \
\STATE{initialize embeddings for the head, relation, and tail entities: $e_h$, $e_r$, and $e_t$}
\STATE{set up the \textit{BI-LSTM} model with initial parameters}
\FOR{\textit{epoch} = 1 to $k$}
    \FOR{each triplet $(h, r, t) \in G$}
        \STATE{obtain \textit{BI-LSTM} outputs: $\tilde{e}_h, \tilde{e}_r, \tilde{e}_t$}
        \STATE{form the concatenated  representation for the triplet as  $q_i$}
    \ENDFOR
    \FOR{each triplet $q_i$}
        \STATE{compute similarity between head neighbor triplets and $q_i$ and obtain $z_{1,i}$}
        \STATE{compute similarity between tail neighbor triplets and $q_i$ and obtain $z_{2,i}$}
    \ENDFOR
    \STATE{compute the loss based on positive and negative samples}
    \STATE{Perform backpropagation and update the model parameters}
\ENDFOR
\RETURN {$q$, $z_1$, $z_2$}
\end{algorithmic}
\end{algorithm}

\subsection{Triplet-view for detecting anomalies}

\begin{figure}[ht]
    \centering
    \includegraphics[clip,scale=0.34]{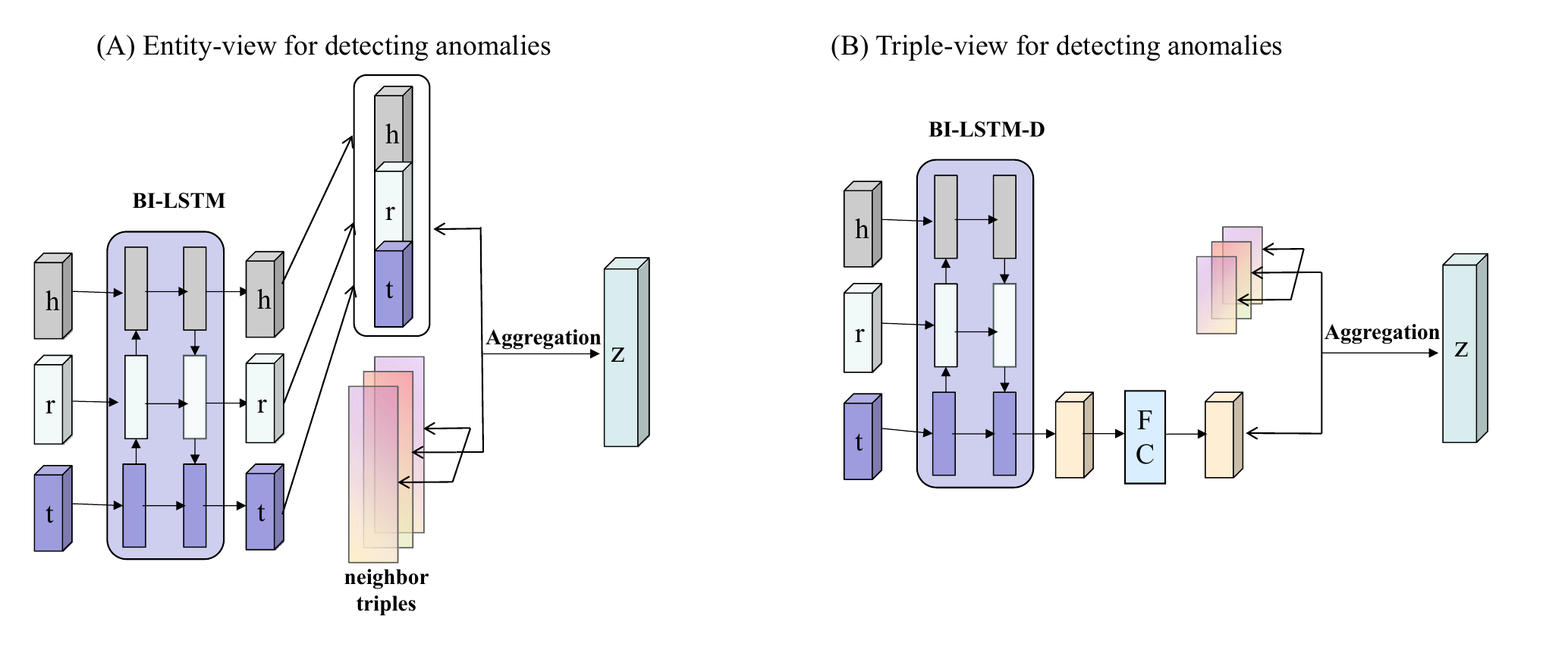}
    \caption{The left is entity-view for detecting anomalies and the right is triplet-view.}
    \label{fig:hebing}
\end{figure}

In this part, as shown in Figure \ref{fig:hebing} (B), we use \textit{BI-LSTM-D} to capture the relationships between entities and relations while reducing the dimensionality. The dimension can be denoted as: $\mathbb{R}^n \times \mathbb{R}^n \times \mathbb{R}^n \rightarrow \mathbb{R}^n$. Formally, given the embeddings of the head entity $e_h$, relation $e_r$, and the tail entity $e_t$, the \textit{BI-LSTM-D} outputs can be expressed as:
\begin{equation}
\tilde{e}_{h,r,t} = \textit{BI-LSTM-D}(e_h, e_r, e_t), \forall (h, r, t) \in G,
\end{equation}
where $G$ represents the set of triplets in the knowledge graph. The \textit{BI-LSTM-D} integrates the information of the entire sequence into the final hidden state, where $\tilde{e}_{h,r,t}$ is the hidden state of the last time step of output. We use this output as the representation for each triplet:
\begin{equation}
q_i' = [\tilde{e}_{h,r,t}].
\end{equation}

Next, we aggregate these representations with their neighbor triplets to derive more robust embeddings. The calculations are similar to the entity-view. The aggregation is performed by weighing the neighbor triplets according to these normalized similarities. This process results in obtaining $z_3$:
\begin{equation}
z_3 = \sum_{j} \textit{softmax}(\textit{sim}(q_i', \textit{q}_j')) \cdot \textit{q}_j',
\end{equation}
Similarly, for each triplet, we compute the similarity with multiple tail neighbor triplets. These similarities are also normalized by using the \textit{softmax} function. The aggregation is done by weighing the neighbor triplets according to the normalized similarities to obtain $z_4$:
\begin{equation}
z_4 = \sum_{k} \textit{softmax}(\textit{sim}(q_i', \textit{q}_k')) \cdot \textit{q}_k',
\end{equation}
where \textit{sim}$(q_i', q_k')$ denotes the similarity between the anchor triplet $q_i'$ and a tail neighbor triplet $q_k'$, also calculated using the dot product.

\begin{algorithm}
    \small
    \caption{triplet-view anomaly detection}
    \label{alg:triplet_view}
    \begin{algorithmic}[1]
    \renewcommand{\algorithmicrequire}{\textbf{Input:}}
    \renewcommand{\algorithmicensure}{\textbf{Output:}}
    \REQUIRE{a knowledge graph $G$; the number of epochs $k$.}
    \ENSURE{node representations: $q'$, $z_3$, and $z_4$.} \
    
\STATE{initialize embeddings for the head, relation, and tail entities: $e_h$, $e_r$, $e_t$}
\STATE{set up the \textit{BI-LSTM-D} model with initial parameters}
\FOR{\textit{epoch} = 1 to $k$}
    \FOR{each triplet $(h, r, t) \in G$}
        \STATE{obtain \textit{BI-LSTM-D} output: $\tilde{e}_{h,r,t}$}
        \STATE{form the representation for the triplet as $q_i'$}
    \ENDFOR
    \FOR{each triplet $q_i'$}
        \STATE{compute similarity between head neighbor triplets and $q_i^{'}$ and obtain $z_{3,i}$}
        \STATE{compute similarity between tail neighbor triplets and $q_i^{'}$ and obtain $z_{4,i}$}
    \ENDFOR
    \STATE{compute the loss based on positive and negative samples}
    \STATE{perform backpropagation and update the model parameters}
\ENDFOR
\RETURN {$q'$, $z_3$, $z_4$}
\end{algorithmic}
\end{algorithm}

\subsection{Training with consistency loss}

\begin{figure}[ht]
    \centering
    \includegraphics[clip,scale=0.4]{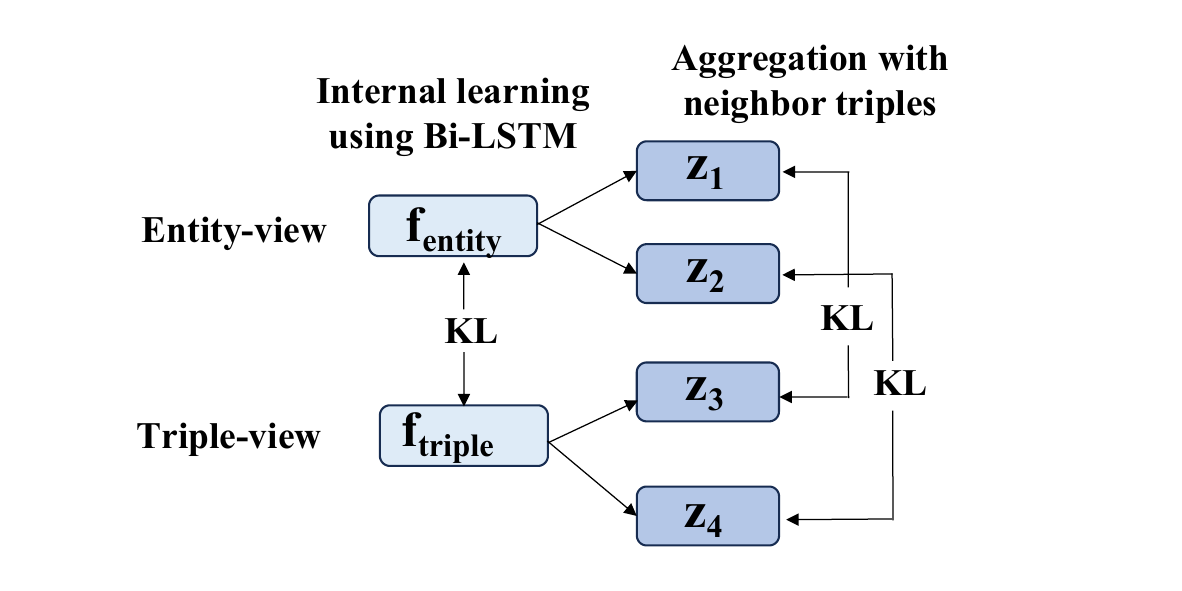}
    \caption{Training with consistency loss.}
    \label{fig:kl}
\end{figure}

In this section, we introduce the consistency loss to ensure that the representations from the two perspectives (entity-view and triplet-view) are consistent with each other. As shown in Figure \ref{fig:kl}, the consistency is enforced through the Kullback-Leibler (KL) divergence. First, we calculate the scores for the two \textit{BI-LSTM} models. The score for the entity-view is calculated by using the norm of the vector difference:
\begin{equation}
f_{\textit{entity}} = \| h + r - t \|_2,
\end{equation}
where $h$, $r$, and $t$ are the embeddings of the head entity, relation, and the tail entity, respectively. This score represents the score of the given triplet $(h, r, t)$ in the context of the entity-view. In addition, the score for the triplet-view is calculated using a multi-layer perceptron ($\textit{MLP}_1$):
\begin{equation}
f_{\textit{triplet}} = \textit{MLP}_1(q_i'),
\end{equation}
where $q_i'$ is the output vector from \textit{BI-LSTM-D}, which integrates the information of the entire sequence into the final hidden state. $MLP_1$ takes this integrated representation and produces a score indicating the score of the triplet in the context of the triplet-view. To ensure consistency between these scores, we compute the KL divergence between the distributions of $f_{\textit{entity}}$ and $f_{\textit{triplet}}$:
\begin{equation}
\mathcal{L}_{\textit{KL, score}} = \textit{KL}(f_{\textit{entity}} \parallel f_{\textit{triplet}}).
\end{equation}
where the KL divergence is calculated by using the following formula, $
\textit{KL}(P \parallel Q) = \sum_{i} P^{(i)} \log \left( \frac{P^{(i)}}{Q^{(i)}} \right),$
where $P^{(i)}$ and $Q^{(i)}$ represent the probability distributions of $f_{\textit{entity}}$ and $f_{\textit{triplet}}$, respectively. In the context of our model, this can be expressed as:
\begin{equation}
\mathcal{L}_{\textit{KL, score}} = \sum_{i} f_{\textit{entity}}^{(i)} \log \left( \frac{f_{\textit{entity}}^{(i)}}{f_{\textit{triplet}}^{(i)}} \right),
\end{equation}
where $f_{\textit{entity}}^{(i)}$ is the probability distribution of the entity-view scores and $f_{\textit{triplet}}^{(i)}$ is the probability distribution of the triplet-view scores. This loss term penalizes the model if the scores from the two views are significantly different, encouraging the model to produce similar scores for the same triplet from both perspectives.

Next, we establish the consistency loss for the aggregation of neighbor triplets. For the head neighbors, the aggregation from the entity-view and triplet-view are scored using different $MLP$. The head neighbor aggregation for the entity-view is scored using $MLP_2$:
\begin{equation}
s_{\textit{head, entity}} = \textit{MLP}_2(z_1),
\end{equation}
where $z_1$ is the aggregated representation of the head neighbors from the entity-view. $MLP_2$ takes this aggregated representation and produces a score. This is the score of the head neighbor aggregation in the entity-view. In addition, the head neighbor aggregation in the triplet-view is scored using $\textit{MLP}_3$:
\begin{equation}
s_{\textit{head, triplet}} = \textit{MLP}_3(z_3),
\end{equation}
where $z_3$ is the aggregated representation of the head neighbors from the triplet-view. $\textit{MLP}_3$ takes this aggregated representation and produces a score. This is a score of the head neighbor aggregation in the triplet-view.
To enforce consistency between these two distributions, we use the KL divergence:
\begin{equation}
\mathcal{L}_{\textit{KL, head}} = \textit{KL}(s_{\textit{head, entity}} \parallel s_{\textit{head, triplet}}).
\end{equation}
This loss term ensures that the scores from the head neighbor aggregations in both views are aligned. This promotes consistency in the head neighbor context representation.

Similarly, the tail neighbors use the same score functions as the head neighbors. The tail neighbor aggregation for the entity-view is scored using $MLP_2$:
\begin{equation}
s_{\textit{tail, entity}} = \textit{MLP}_2(z_2),
\end{equation}
where $z_2$ is the aggregated representation of the tail neighbors from the entity-view. $\textit{MLP}_2$ produces a score for the tail neighbors' context in the entity-view. In addition, the tail neighbor aggregation in the triplet-view is scored using $\textit{MLP}_3$:
\begin{equation}
s_{\textit{tail, triplet}} = \textit{MLP}_3(z_4),
\end{equation}
where $z_4$ is the aggregated representation of the tail neighbors from the triplet-view. $\textit{MLP}_3$ produces a score for the tail neighbors' context in the triplet-view. We also compute the KL divergence between these two distributions to enforce similarity:
\begin{equation}
\mathcal{L}_{\textit{KL, tail}} = \textit{KL}(s_{\textit{tail, entity}} \parallel s_{\textit{tail, triplet}}).
\end{equation}
This loss term ensures that the scores from the tail neighbor aggregations in both views are aligned, promoting consistency in the tail neighbor context representation.

The final consistency loss is the sum of the KL divergence losses, i.e.,
\begin{equation}
\mathcal{L}_{\textit{consistency}} = \mathcal{L}_{\textit{KL, score}} + \mathcal{L}_{\textit{KL, head}} + \mathcal{L}_{\textit{KL, tail}}.
\end{equation}
This comprehensive consistency loss ensures that the embeddings and their aggregations from both perspectives are aligned, promoting robustness and coherence in the representations learning.

\subsection{Detecting anomalies}

The final training loss for detecting anomalies is composed of two main parts. The first part of the loss function focuses on ensuring that the positive samples have a lower score than the negative samples by a specified margin. This is achieved by the following loss calculation:
\begin{equation} \label{margin}
\mathcal{L}_{\textit{margin}} = \textit{margin}(\textit{loss}_p, \textit{loss}_n) = \max(0, \textit{loss}_n - \textit{loss}_p + \gamma),
\end{equation}
where $\textit{loss}_p$ represents the sum of the scores for all positive samples and $\textit{loss}_n$ represents the sum of the scores for all negative samples, and  $\gamma$ is the margin parameter. This method emphasizes that the scores for positive samples should be at least one margin lower than the scores for negative samples, thus distinguishing correct triplets from anomalies.

Each sample's $\textit{loss}_p$ consists of two parts: the score from the \textit{BI-LSTM} layer and the similarity scores from the aggregation with neighbor triplets. The combined score for each sample can be expressed as:
\begin{equation}
\textit{loss}_p = \alpha \cdot f_{\textit{BI}}  + (1 - \alpha) \cdot \left( \frac{1}{2} \left( \textit{sim}(z_1, z_2) + \textit{sim}(z_3, z_4) \right) \right),
\end{equation}
where $f_{\textit{BI}}$ is the score from the \textit{BI-LSTM} layer. In entity-view, it is calculated by $||h+r-t||_{2}$, and in the triplet-view, it is calculated with $MLP_1(\cdot)$, as followings:
$f_{\textit{BI}}$ = $f_{\textit{BI-LSTM}}$ + $f_{\textit{BI-LSTM-D}}$  = $||\tilde{e}_h+ \tilde{e}_r- \tilde{e}_t||_{2} + MLP_1(\tilde{e}_{h,r,t})$. In addition, $\textit{sim}(z_1, z_2)$ is the similarity between the head neighbor triplets aggregation and tail neighbor triplets aggregation in the entity-view, and $\textit{sim}(z_3, z_4)$ is the similarity between the head neighbor triplets aggregation and tail neighbor triplets aggregation in the triplet-view. The similarity function $\textit{sim}$ is calculated by the dot product, which is defined as $\textit{sim}(a, b)$ = $a \cdot b$. The parameter $\alpha$ is a weight that balances the contributions between internal information aggregation and neighbor triplets aggregation. The calculation for negative samples is similar to the positive sample.

The second part of the training loss is the consistency loss, which ensures that the representations from the two perspectives (entity-view and triplet-view) are similar. This is achieved by calculating the kullback-leibler (KL) divergence between the scores from the two views. The consistency loss is defined as:
\begin{equation}
\mathcal{L}_{\textit{consistency}} = \mathcal{L}_{\textit{KL, score}} + \mathcal{L}_{\textit{KL, head}} + \mathcal{L}_{\textit{KL, tail}},
\end{equation}

By combining these two parts, the total training loss can be expressed as:
\begin{equation}
\mathcal{L} = (1-\beta) \cdot \mathcal{L}_{\textit{margin}} + \beta \cdot\mathcal{L}_{\textit{consistency}}.
\end{equation}

This combined loss function ensures that the model effectively distinguishes between correct triplets and anomalies while maintaining consistency between the entity-view and triplet-view representations.
Finally, we use the best prediction model to complete the prediction. The combined score for each sample is calculated as follows:
\begin{equation}
\textit{score} = \alpha \cdot (f_{\textit{BI-LSTM}} + f_{\textit{BI-LSTM-D}}) + (1 - \alpha) \cdot \left( \frac{1}{2} \left( \textit{sim}(z_1, z_2) + \textit{sim}(z_3, z_4) \right) \right).
\end{equation}
In the final prediction, it does not need to use the negative samples or KL divergence. As a result, a higher score indicates a higher likelihood of the sample being an anomaly.

\subsection{Time complexity}

The time complexity of the ADKGD framework can be divided into three main stages: \textbf{data preparation}, \textbf{training}, and \textbf{inference}. Each stage involves specific computational steps, as detailed below.
During \textbf{data preparation}, negative sampling and neighbor extraction are the main operations. Negative sampling involves generating $ N $ negative samples for each triplet in the knowledge graph ($ G $). For a total of $ |T| $ triplets, the complexity of this step is $ O(|T| \cdot N) $. Additionally, extracting head and tail neighbor triplets requires searching for triplets sharing the same head or tail entity. Given an average entity degree of $ d $, the complexity for neighbor extraction is $ O(|T| \cdot d) $.
The \textbf{training phase} consists of representation learning, neighbor aggregation, score computation, consistency loss calculation, and backpropagation. For representation learning, the \textit{BI-LSTM} processes each triplet, resulting in a complexity of $ O(|T| \cdot n^2) $, where $ n $ is the embedding dimensionality. Similarly, the \textit{BI-LSTM-D} operates on reduced dimensions with a complexity of $ O(|T| \cdot n) $. Neighbor aggregation involves calculating similarities between a triplet and its neighbors. With $ k $ average neighbors per triplet, the complexity of this operation is $ O(|T| \cdot k \cdot n) $. The score computation step requires $ O(|T| \cdot n^2) $, and the consistency loss, which enforces alignment between the entity-view and triplet-view, has a complexity of $ O(|T| \cdot n) $. Backpropagation, which updates the model parameters, depends on the number of parameters ($ p $) and has a complexity of $ O(p) $. Combining these components, the total time complexity for training per epoch is $ O(|T| \cdot (n^2 + k \cdot n)) + O(p) $.
In the \textbf{inference phase}, the model computes scores for all triplets based on the trained representations. Score computation, similar to training, has a complexity of $ O(|T| \cdot n^2) $. Neighbor aggregation, which processes the context for each triplet, incurs an additional complexity of $ O(|T| \cdot k \cdot n) $. Thus, the total complexity for inference is $ O(|T| \cdot (n^2 + k \cdot n)) $.
In summary, the overall time complexity for the three stages of ADKGD is as follows: data preparation has a complexity of $ O(|T| \cdot (N + d)) $, training per epoch has a complexity of $ O(|T| \cdot (n^2 + k \cdot n)) + O(p) $, and inference has a complexity of $ O(|T| \cdot (n^2 + k \cdot n)) $. Among these, the dominant term in training and inference is $ O(|T| \cdot n^2) $, highlighting that embedding dimensionality ($ n $) significantly impacts computational efficiency. Optimizing $ n $ or reducing $ k $ through more efficient neighbor aggregation strategies could improve scalability for large-scale knowledge graphs.

\section{Experiments} \label{sec: experiments}

Our experiments were carried out on a system equipped with a single NVIDIA GeForce RTX 3090 Ti GPU. Detailed information about the experimental setup is provided below. In this section, we assess ADKGD, aiming to address the following research questions:

\begin{itemize}
        \item \textbf{Q1 (Effectiveness):} How effective is ADKGD compared with the state-of-the-art KG anomaly detection methods?
        
        \item \textbf{Q2 (Ablation study):} How does each component of ADKGD contribute to its performance?
        
        \item \textbf{Q3 (Parameter analysis):} How do the hyperparameters influence the performance of ADKGD?
    
        \item \textbf{Q4 (Time  effciency):} How does the computational time of ADKGD?
\end{itemize}

\subsection{Datasets and evaluation metrics}

We conduct comprehensive experiments on several real KGs: FB15K-237, WN18RR, NELL-995, Kinship, Yago, and KG20C.

\textbf{FB15K-237 \cite{toutanova2015representing}}: FB15K-237 is a challenging subset of Freebase with 14,541 entities and 237 relations, derived by removing inverse relations from FB15K to prevent data leakage.

\textbf{WN18RR \cite{dettmers2018convolutional}}: WN18RR is a subset of WordNet, featuring 40,943 entities and 11 relations. It retains the hierarchical structure of WordNet. Additionally, it addresses the test leakage issues found in the original WN18 dataset.

\textbf{NELL-995 \cite{balazevic2019multi}}: NELL-995 is a subset of the never-ending language learner (NELL) with 200 relations. It includes specifically created splits such as NELL-995-h25, NELL-995-h50, NELL-995-h75, and NELL-995-h100. These splits contain various hierarchical relations to evaluate the impact of hierarchical structures on model performance.

\textbf{Kinship \cite{robinson2016families}}: Kinship is a dataset of family relations, containing a total of 104 entities and 46 relations, with 6,529 triplets. It is commonly used for evaluating knowledge graph models in the context of familial relationships.

\textbf{Yago \cite{yago}}: YAGO is a large-scale knowledge graph derived from Wikipedia, WordNet, and other sources, containing 123,182 entities and 37 relations. It is one of the largest publicly available KGs, widely used for various knowledge graph tasks.

\textbf{KG20C \cite{tran_exploringscholarlydata_2019}}: KG20C is a large-scale scholarly knowledge graph that represents academic research relationships, including 16,362 entities and 5 relations. It focuses on academic papers, authors, and their citation relationships.

These datasets are used to evaluate the performance of ADKGD across different types of KGs, ensuring the robustness and generalizability of the results. Following the previous studies \cite{jia2019triple, shan2018confidence, zhang2022contrastive}, we employ three real-world datasets constructed with noisy triplets. These noisy triplets are included at Ratio of 5\%, 10\%, and 15\% of the entire knowledge graphs, based on popular benchmarks. The statistical information of these datasets is summarized in Table \ref{tab:datasets}.

\begin{table}[ht]
    \small
    \centering
    \caption{The statistical information of the datasets.}
    \label{tab:datasets}
    \begin{tabular}{lcccc}
    \hline
    \textbf{Dataset} & \textbf{Entities} & \textbf{Relations} & \textbf{triplets} & \textbf{Mean in-degree} \\
    \hline
    FB15K-237 & 14,541 & 237 & 310,116 & 22.96 \\
    WN18RR & 40,943 & 11 & 93,003 & 2.12 \\
    NELL-995-h25 & 70,145 & 172 & 140,999 & 4.18 \\
    NELL-995-h50 & 34,667 & 86 & 83,600 & 5.00 \\
    NELL-995-h75 & 28,085 & 57 & 67,965 & 4.43 \\
    NELL-995-h100 & 22,411 & 43 & 57,823 & 4.73 \\
     Kinship & 104 & 46 & 6,529  &  62.85 \\
    Yago & 123,182 & 37 & 1,089,040 & 8.84 \\
    KG20C & 16,362 & 5 & 55,607 & 3.40 \\
    \hline
    \end{tabular}
\end{table}

To evaluate the performance of all the compared approaches, we use ranking measures similar to those used in the benchmark methods \cite{zhang2022contrastive}.  Specifically, we rank all the triplets in the target KG according to their anomalies score in descending order. A triplet with a higher score would be more likely to be an anomaly. To fairly evaluate the performance, we use the following two evaluation measures:

Precision@\textbf{K} represents the proportion of real false triplets among those with the top $K$ highest scores. It is calculated as:
\begin{equation}
Precision@\textbf{K} = \frac{|\textit{Errors Discovered in Top K Ranking List}|}{|K|}    
\end{equation}

Recall@\textbf{K} denotes the proportion of real false triplets with the top $K$ highest scores relative to the \begin{equation}
Recall@\textbf{K} = \frac{|\textit{Errors Discovered in Top K Ranking List}|}{|\textit{Total Number of Errors in KG}|}
\end{equation}

These metrics allow us to effectively measure the ability of the models to identify anomaly triplets in the kG.

\subsection{Baselines and experiment setting}

In the experiments, we include two categories of baselines. First, KG-embedding based methods, including \textbf{TransE} \cite{bordes2013translating}, \textbf{DistMult} \cite{yang2015embedding}, and \textbf{ComplEx} \cite{trouillon2016complex}. To perform anomaly detection, after learning the embedding representations, we assess the triplets based on the corresponding score functions. Second, we include some state-of-the-art KG anomaly detection methods: \textbf{CKRL} \cite{shan2018confidence}, \textbf{KGTtm} \cite{jia2019triple}, \textbf{KGIst} \cite{belth2020normal}, \textbf{CAGED} \cite{zhang2022contrastive}. \textbf{CKRL} enhances TransE by considering all possible paths between the head entity and the tail entity. \textbf{KGTtm} further enhances CKRL by integrating the global graph structure of the KG. \textbf{KGIst} proposes an unsupervised method to learn soft rules and detect errors based on these rules. \textbf{CAGED} learns the scoring function with cross-layer, including internal information aggregation with \textit{BI-LSTM} and aggregation with neighbor triplets. In addition, we compare the text-based method \textbf{CCA} \cite{liu2024knowledge} and \textbf{SeSICL} \cite{liu2024sesicl}.

In our ADKGD model, we adopt a similar experiment setting as CAGED \cite{zhang2022contrastive}. Firstly, we optimize all models using the Adam optimizer with a fixed batch size of 256. The default Xavier initializer is used to initialize model parameters, and the initial learning rate is set to 0.01. The embedding size is fixed at 100 for all models. We apply a grid search for hyperparameter tuning. The margin parameter $\gamma$ from 0 to 1,  the trade-off parameters $\alpha$ between 0.1 and 0.9 and $\beta$ between 0.1 and 0.9. In addition, the number of neighbors is computed by taking the average number of neighbors of all triplets in a dataset. This ensures that our model can adapt to datasets with different densities of neighbors, thereby making the best use of neighborhood information. To reduce randomness, we use a fixed random seed and report the average results over ten runs.


\subsection{Effectiveness of ADKGD (Q1)}

To answer Q1, we conduct comprehensive experiments on three real-world KGs. The experimental results with an anomaly ratio equal to 5\% are summarized in Table \ref{table:precision_recall}. It is evident from Table \ref{table:precision_recall} that KG anomaly detection methods such as ADKGD, CAGED, CKRL, KGTtm, and KGIst outperform KG-embedding based methods like TransE, ComplEx, and DistMult. This superior performance can be attributed to the fact that traditional KG embedding frameworks do not account for the presence of errors in the KG. Consequently, they fail to learn discriminative representations that can differentiate between normal and noisy triplets. 

Among all the evaluated methods, ADKGD consistently achieves the highest precision and recall across all datasets and values of K. For instance, at K = 5\%, the precision for ADKGD on the FB15K dataset is 0.659, significantly outperforming the average precision of KG-embedding-based methods, which is 0.484, representing an improvement of 36.2\%. Similarly, for the WN18RR dataset at K = 5\%, ADKGD achieves a precision of 0.560, compared to the average precision of 0.320 for KG-embedding-based methods, showing an increase of 24\%. In the NELL-995 dataset, ADKGD attains a precision of 0.976 at K = 1\%, which is a 33.8\% improvement over the average precision of 0.638 for KG-embedding-based methods at the same K value. Additionally, ADKGD maintains a high precision of 0.650 even at K = 5\%, well above the average precision of 0.385 for KG-embedding-based methods, representing an increase of 26.5\%.

When comparing ADKGD to state-of-the-art KG anomaly detection methods, the superiority of ADKGD remains evident. First, let’s consider the average performance of these methods. For the FB15K dataset at K = 1\%, the average precision of CKRL, KGTtm, KGIst, and CAGED is 0.839. At K = 5\%, the average precision is 0.617. ADKGD surpasses this average with a precision of 0.951 at K = 1\%, representing a 13.3\% improvement, and 0.659 at K = 5\%, representing a 6.8\% improvement. CAGED is the top performer among the state-of-the-art KG anomaly detection methods. Thus, we conduct a detailed comparison between ADKGD and CAGED. In the FB15K dataset, at K = 1\%, CAGED achieves a precision of 0.927, whereas ADKGD excels with an even higher precision of 0.951, representing an improvement of 2.6\%. This trend continues at K = 5\%, where ADKGD attains a precision of 0.659 compared to CAGED's 0.656, showing a modest increase of 0.5\%. For the WN18RR dataset, ADKGD maintains its superiority with a precision of 0.943 at K = 1\%, compared to CAGED’s 0.847, which is an 11.3\% improvement. At K = 5\%, ADKGD achieves a precision of 0.560, surpassing CAGED’s 0.482 by 16.2\%. In the NELL-995 dataset, ADKGD achieves an outstanding precision of 0.976 at K = 1\%, which is 2.2\% higher than CAGED's 0.955. It also maintains a high precision of 0.650 at K = 5\%, representing a 5.9\% increase over CAGED's 0.614.

In addition, we compared ADKGD with the CAGED, conducting anomaly detection experiments on three additional datasets: Kinship, Yago, and KG20C. We analyze the performance of ADKGD on these datasets. ADKGD demonstrates strong anomaly detection capabilities across the Kinship, Yago, and KG20C datasets, showcasing its effectiveness in graph-based anomaly detection tasks. On the Kinship dataset, ADKGD achieves high Precision@K at low K values (e.g., K = 1\% and K = 2\%), outperforming other graph-based methods. Overall, ADKGD's performance on these three datasets confirms its effectiveness in KG-based anomaly detection tasks. It excels at identifying a small number of anomalies with high accuracy, particularly at low K values. However, when the number of anomalies increases, text-based methods (e.g., SeSICL and CCA) show a clear advantage, especially in achieving higher recall at larger K values by capturing hidden anomaly patterns.
At K = 5\%, CCA slightly surpasses ADKGD, indicating that incorporating textual data can help capture more anomalies when a larger number is considered. Similarly, on the Yago dataset, ADKGD performs well at lower K values but falls behind SeSICL and CCA in terms of Recall@K at K = 5\%, highlighting the advantage of text-based methods in capturing a broader range of anomalies. Lastly, on the KG20C dataset, ADKGD achieves the best Recall@K performance, particularly at K = 5\%, detecting more anomalies overall. However, CCA continues to perform better in Precision@K, demonstrating the advantage of textual information in precisely identifying anomalies. Although ADKGD demonstrates excellent anomaly detection performance without using textual data, it still falls short compared to text-based methods such as SeSICL and CCA. Textual information provides additional semantic context, helping detect more anomalies that are difficult to identify solely based on graph structures. The results in the table show that text-based methods achieve better Recall@K at higher K values, indicating their superiority in handling tasks that require detecting a large number of anomalies.
Currently, ADKGD relies only on the structural information of knowledge graphs for anomaly detection, without leveraging external textual data. As a result, it underperforms in recall and precision at higher K values compared to methods that incorporate textual information, such as SeSICL and CCA. On datasets such as Kinship, Yago, and KG20C, text-based methods utilize external textual information to capture broader anomaly patterns, especially when more anomalies need to be detected. In contrast, ADKGD is limited by its reliance on graph structure, which restricts its ability to detect all potential anomalies effectively. Therefore, future work will focus on integrating textual data with graph structure to enhance the anomaly detection capabilities of ADKGD. By incorporating external textual information, we believe ADKGD can maintain its advantages in graph-based anomaly detection while further improving recall and precision, particularly at higher K values. This hybrid approach of combining graph and textual data has the potential to achieve even better performance in anomaly detection tasks.

The experimental results indicate that anomaly detection methods designed explicitly for KGs are more effective than traditional KG embedding methods. Among these, ADKGD stands out as particularly effective. In addition, it performs better than other state-of-the-art KG anomaly detection methods. This is because ADKGD allows for more comprehensive feature extraction and better integration of information across different layers and channels, leading to more accurate anomaly detection. The incorporation of cross-layer and dual-channel training significantly enhances the performance of these methods in anomaly detection in KGs.

\begin{table}[ht]
\centering
\caption{Anomaly detection results of Precision@K and Recall@K based on the six datasets with anomaly ratio = 5\%.}
\label{table:precision_recall}
\resizebox{\textwidth}{!}{%
\begin{tabular}{lccccccccccccccc}
\toprule
 & \multicolumn{5}{c}{FB15K} & \multicolumn{5}{c}{WN18RR} & \multicolumn{5}{c}{NELL-995-h25} \\
 & $K$ = 1\% & $K$ = 2\% & $K$ = 3\% & $K$ = 4\% & $K$ = 5\% & $K$ = 1\% & $K$ = 2\% & $K$ = 3\% & $K$ = 4\% & $K$ = 5\% & $K$ = 1\% & $K$ = 2\% & $K$ = 3\% & $K$ = 4\% & $K$ = 5\% \\
\midrule
\multicolumn{16}{c}{Precision@K} \\
\midrule
TransE & 0.756 & 0.674 & 0.605 & 0.546 & 0.485 & 0.581 & 0.488 & 0.371 & 0.345 & 0.331 & 0.659 & 0.550 & 0.476 & 0.423 & 0.383 \\
ComplEx & 0.718 & 0.651 & 0.590 & 0.534 & 0.485 & 0.518 & 0.444 & 0.382 & 0.341 & 0.307 & 0.627 & 0.538 & 0.472 & 0.427 & 0.373 \\
DistMult & 0.709 & 0.646 & 0.582 & 0.529 & 0.483 & 0.574 & 0.451 & 0.390 & 0.349 & 0.322 & 0.630 & 0.553 & 0.493 & 0.446 & 0.400 \\
CKRL & 0.789 & 0.736 & 0.684 & 0.630 & 0.579 & 0.675 & 0.542 & 0.456 & 0.389 & 0.349 & 0.735 & 0.642 & 0.559 & 0.498 & 0.450 \\
KGttm & 0.815 & 0.767 & 0.713 & 0.612 & 0.579 & 0.770 & 0.628 & 0.516 & 0.444 & 0.396 & 0.808 & 0.691 & 0.602 & 0.535 & 0.481 \\
KGIst & 0.825 & 0.754 & 0.703 & 0.617 & 0.569 & 0.747 & 0.599 & 0.476 & 0.407 & 0.379 & 0.782 & 0.678 & 0.584 & 0.528 & 0.485 \\
CAGED & 0.927 & 0.867 & 0.798 & 0.729 & 0.656 & 0.847 & 0.702 & 0.608 & 0.537 & 0.482 & 0.955  &  0.869 & 0.779 & 0.686 & 0.614 \\
\textbf{ADKGD} & \textbf{0.951} & \textbf{0.885} & \textbf{0.812} & \textbf{0.734} & \textbf{0.659} & \textbf{0.943} & \textbf{0.845} & \textbf{0.724} & \textbf{0.635} & \textbf{0.560} & \textbf{0.976} & \textbf{0.917} & \textbf{0.835} & \textbf{0.734} & \textbf{0.650} \\
\midrule
SeSICL & 0.963 & 0.897 & 0.845 & 0.759 & 0.696 & 0.953 & 0.867 & 0.756 & 0.681 & 0.612 & 0.985  &  0.931 & 0.850 & 0.751 & 0.667 \\
CCA & \textbf{0.964} & \textbf{0.904} & \textbf{0.853} & \textbf{0.807} & \textbf{0.709} & \textbf{0.967} & \textbf{0.869} & \textbf{0.816} & \textbf{0.767} & \textbf{0.660} & \textbf{0.986} & \textbf{0.947} & \textbf{0.885} & \textbf{0.794} & \textbf{0.723} \\
\midrule
\multicolumn{16}{c}{Recall@K} \\
\midrule
TransE & 0.151 & 0.270 & 0.363 & 0.437 & 0.488 & 0.116 & 0.195 & 0.223 & 0.276 & 0.331 & 0.132 & 0.220 & 0.285 & 0.338 & 0.383 \\
ComplEx & 0.143 & 0.260 & 0.354 & 0.427 & 0.485 & 0.103 & 0.177 & 0.229 & 0.273 & 0.307 & 0.125 & 0.215 & 0.283 & 0.341 & 0.373 \\
DistMult & 0.141 & 0.258 & 0.349 & 0.423 & 0.483 & 0.114 & 0.180 & 0.233 & 0.279 & 0.322 & 0.126 & 0.216 & 0.290 & 0.347 & 0.400 \\
CKRL & 0.150 & 0.294 & 0.411 & 0.504 & 0.579 & 0.120 & 0.198 & 0.277 & 0319 & 0.349 & 0.139 & 0.239 & 0.329 & 0.387 & 0.450 \\
KGttm & 0.163 & 0.307 & 0.428 & 0.490 & 0.579 & 0.154 & 0.251 & 0.309 & 0.355 & 0.396 & 0.149 & 0.256 & 0.342 & 0.412 & 0.481 \\
KGIst & 0.165 & 0.302 & 0.428 & 0.490 & 0.569 & 0.147 & 0.233 & 0.295 & 0.341 & 0.379 & 0.148 & 0.252 & 0.341 & 0.406 & 0.485 \\
CAGED & 0.185 & 0.347 & 0.479 & 0.583 & 0.656 & 0.169 & 0.281 & 0.364 & 0.430 & 0.482 & 0.190 & 0.347 & 0.467 & 0.549 & 0.614 \\
\textbf{ADKGD} & \textbf{0.190} & \textbf{0.354} & \textbf{0.487} & \textbf{0.587}  & \textbf{0.659} & \textbf{0.188} & \textbf{0.338} & \textbf{0.434} & \textbf{0.508} & \textbf{0.560} & \textbf{0.195} & \textbf{0.367} & \textbf{0.501} & \textbf{0.587} & \textbf{0.650} \\
\midrule
SeSICL & 0.205 & 0.370 & 0.517 & 0.613 & 0.696 & 0.213 & 0.359 & 0.457 & 0.524 & 0.612 & 0.205 & 0.375 & 0.510 & 0.600 & 0.667 \\
\textbf{CCA} & 0.218 & 0.381 & 0.527 & 0.625 & 0.709 & 0.216 & 0.368 & 0.469 & 0.534 & 0.660 & 0.215 & 0.385 & 0.515 & 0.605 & 0.723 \\
\bottomrule
\toprule
 & \multicolumn{5}{c}{Kinship} & \multicolumn{5}{c}{Yago} & \multicolumn{5}{c}{KG20C} \\
 & $K$ = 1\% & $K$ = 2\% & $K$ = 3\% & $K$ = 4\% & $K$ = 5\% & $K$ = 1\% & $K$ = 2\% & $K$ = 3\% & $K$ = 4\% & $K$ = 5\% & $K$ = 1\% & $K$ = 2\% & $K$ = 3\% & $K$ = 4\% & $K$ = 5\% \\
\midrule
\multicolumn{16}{c}{Precision@K} \\
\midrule
TransE & 0.720 & 0.625 & 0.563 & 0.505 & 0.450 & 0.553 & 0.470 & 0.355 & 0.330 & 0.310 & 0.620 & 0.510 & 0.440 & 0.390 & 0.360 \\
ComplEx & 0.685 & 0.615 & 0.563 & 0.510 & 0.465 & 0.495 & 0.425 & 0.375 & 0.335 & 0.305 & 0.600 & 0.510 & 0.445 & 0.400 & 0.355 \\
DistMult & 0.675 & 0.605 & 0.545 & 0.495 & 0.460 & 0.555 & 0.435 & 0.380 & 0.340 & 0.310 & 0.595 & 0.515 & 0.465 & 0.420 & 0.380 \\
CKRL & 0.750 & 0.695 & 0.640 & 0.585 & 0.535 & 0.625 & 0.495 & 0.410 & 0.350 & 0.315 & 0.715 & 0.625 & 0.540 & 0.485 & 0.435 \\
KGttm & 0.790 & 0.745 & 0.685 & 0.590 & 0.560 & 0.740 & 0.600 & 0.490 & 0.420 & 0.380 & 0.800 & 0.685 & 0.595 & 0.530 & 0.475 \\
KGIst & 0.805 & 0.715 & 0.655 & 0.570 & 0.525 & 0.725 & 0.580 & 0.460 & 0.395 & 0.365 & 0.765 & 0.670 & 0.575 & 0.515 & 0.475 \\
CAGED & 0.810 & 0.740 & 0.670 & 0.605 & 0.545 & 0.740 & 0.635 & 0.550 & 0.480 & 0.420 & 0.860 & 0.780 & 0.690 & 0.615 & 0.540 \\
\textbf{ADKGD} & \textbf{0.849} & \textbf{0.723} & \textbf{0.643} & \textbf{0.555} & \textbf{0.659} & \textbf{0.967} & \textbf{0.920} & \textbf{0.849} & \textbf{0.767} & \textbf{0.687} & \textbf{0.931} & \textbf{0.836} & \textbf{0.719} & \textbf{0.619} & \textbf{0.545} \\
\midrule
SeSICL & 0.860 & 0.770 & 0.690 & 0.610 & 0.675 & 0.920 & 0.870 & 0.810 & 0.740 & 0.670 & 0.965 & 0.890 & 0.805 & 0.725 & 0.655 \\
CCA & \textbf{0.875} & \textbf{0.785} & \textbf{0.710} & \textbf{0.630} & \textbf{0.695} & \textbf{0.943} & \textbf{0.885} & \textbf{0.820} & \textbf{0.764} & \textbf{0.694} & \textbf{0.981} & \textbf{0.912} & \textbf{0.825} & \textbf{0.743} & \textbf{0.678} \\
\midrule
\multicolumn{16}{c}{Recall@K} \\
\midrule
TransE & 0.141 & 0.259 & 0.351 & 0.426 & 0.488 & 0.108 & 0.184 & 0.213 & 0.265 & 0.324 & 0.128 & 0.215 & 0.280 & 0.334 & 0.378 \\
ComplEx & 0.133 & 0.250 & 0.340 & 0.419 & 0.480 & 0.098 & 0.169 & 0.222 & 0.266 & 0.301 & 0.121 & 0.210 & 0.277 & 0.335 & 0.366 \\
DistMult & 0.130 & 0.248 & 0.338 & 0.413 & 0.480 & 0.110 & 0.173 & 0.226 & 0.272 & 0.314 & 0.123 & 0.213 & 0.285 & 0.341 & 0.396 \\
CKRL & 0.140 & 0.280 & 0.396 & 0.487 & 0.559 & 0.115 & 0.190 & 0.267 & 0.310 & 0.338 & 0.130 & 0.230 & 0.318 & 0.375 & 0.438 \\
KGttm & 0.153 & 0.294 & 0.412 & 0.475 & 0.559 & 0.149 & 0.240 & 0.296 & 0.342 & 0.381 & 0.145 & 0.252 & 0.336 & 0.405 & 0.475 \\
KGIst & 0.155 & 0.295 & 0.410 & 0.472 & 0.559 & 0.141 & 0.224 & 0.284 & 0.329 & 0.366 & 0.145 & 0.247 & 0.336 & 0.402 & 0.475 \\
CAGED & 0.155 & 0.290 & 0.395 & 0.455 & 0.515 & 0.170 & 0.280 & 0.350 & 0.410 & 0.465 & 0.170 & 0.300 & 0.410 & 0.485 & 0.550 \\
\textbf{ADKGD} & \textbf{0.168} & \textbf{0.288} & \textbf{0.385} & \textbf{0.443} & \textbf{0.500} & \textbf{0.193} & \textbf{0.368} & \textbf{0.509} & \textbf{0.614} & \textbf{0.687} & \textbf{0.186} & \textbf{0.334} & \textbf{0.431} & \textbf{0.495} & \textbf{0.545} \\
\midrule
SeSICL & 0.180 & 0.300 & 0.395 & 0.455 & 0.675 & 0.200 & 0.380 & 0.520 & 0.625 & 0.670 & 0.195 & 0.340 & 0.440 & 0.510 & 0.655 \\
CCA & \textbf{0.195} & \textbf{0.315} & \textbf{0.405} & \textbf{0.465} & \textbf{0.695} & \textbf{0.211} & \textbf{0.393} & \textbf{0.537} & \textbf{0.635} & \textbf{0.694} & \textbf{0.207} & \textbf{0.358} & \textbf{0.457} & \textbf{0.523} & \textbf{0.678} \\
\bottomrule

\end{tabular}
}
\end{table}

Table \ref{table:others} shows the performance results of ADKGD on different anomaly ratios (5\%, 10\%, and 15\%) for the NELL-995 dataset variants (h25, h50, h75, h100). Across all four NELL-995 dataset variants, the average precision of ADKGD improves as the anomaly ratio increases. Specifically, when the anomaly ratio increases from 5\% to 10\%, the average precision of ADKGD improves by about 25\% across different K values. Further, when the anomaly ratio increases from 10\% to 15\%, the average precision of ADKGD improves by about 10\%. For example, at K = 5\%, the average precision of ADKGD is 0.662 at a 5\% anomaly ratio, which increases to 0.908 at a 10\% anomaly ratio and further increases to 0.965 at a 15\% anomaly ratio. The recall of ADKGD shows a slight decreasing trend as the anomaly ratio increases. When the anomaly ratio increases from 5\% to 10\%, the average recall of ADKGD decreases by about 30\% across different K values. When the anomaly ratio increases from 10\% to 15\%, the average recall of ADKGD further decreases by about 20\%. For example, at K = 5\%, the average recall of ADKGD is 0.662 at a 5\% anomaly ratio, which decreases to 0.453 at a 10\% anomaly ratio and further decreases to 0.321 at a 15\% anomaly ratio. In summary, the precision of ADKGD generally improves with higher anomaly ratios, indicating its robustness and effectiveness in anomaly detection under increased noise levels.

\begin{table}[h]
\centering
\caption{anomaly detection results on variant NELL-995 with different anomaly ratios.}
\label{table:error_detection}
\resizebox{\textwidth}{!}{%
\begin{tabular}{c|c|ccccc|ccccc|ccccc}
\toprule
& \textbf{Ratio} & \multicolumn{5}{c|}{5\%} & \multicolumn{5}{c|}{10\%} & \multicolumn{5}{c}{15\%} \\
\textbf{} & \textbf{K} & K=1\% & K=2\% & K=3\% & K=4\% & K=5\% & K=1\% & K=2\% & K=3\% & K=4\% & K=5\% & K=1\% & K=2\% & K=3\% & K=4\% & K=5\% \\
\midrule
\multirow{4}{*}{\rotatebox{90}{\textbf{Precision}}} & \textbf{NELL-995-h25} & 0.976 & 0.917 & 0.835 & 0.734 & 0.650 & 0.992 & 0.985 & 0.968 & 0.943 & 0.908 & 0.997 & 0.987 & 0.983 & 0.975 & 0.961 \\
& \textbf{NELL-995-h50} & 0.984 & 0.933 & 0.844 & 0.744 & 0.657 & 0.994 & 0.979 & 0.955 & 0.927 & 0.889 & 0.996 & 0.988 & 0.980 & 0.969 & 0.956 \\
& \textbf{NELL-995-h75} & 0.982 & 0.910 & 0.833 & 0.735 & 0.658 & 0.992 & 0.978 & 0.958 & 0.930 & 0.900 & 0.998 &  0.994 & 0.989 & 0.984 & 0.978 \\
& \textbf{NELL-995-h100} &  0.975 & 0.942 & 0.863 &  0.772 & 0.683 & 0.996 & 0.990 & 0.973 & 0.955 & 0.926 & 0.989 &  0.985 & 0.985 & 0.978 & 0.967 \\
\midrule
\multirow{4}{*}{\rotatebox{90}{\textbf{Recall@K}}} & \textbf{NELL-995-h25} & 0.195 & 0.367 & 0.501 & 0.587 & 0.650 & 0.099 & 0.196 & 0.290 & 0.377 & 0.454 & 0.066 & 0.131 & 0.196 & 0.260 & 0.320 \\
& \textbf{NELL-995-h50} & 0.196 & 0.373 & 0.506 & 0.595 & 0.657 & 0.099 & 0.195 & 0.286 & 0.370 & 0.444 & 0.066 & 0.131 & 0.196 & 0.258 & 0.318 \\
& \textbf{NELL-995-h75} & 0.196 & 0.364 & 0.499 & 0.587 & 0.658 & 0.099 & 0.195 & 0.287 & 0.371 & 0.450 &  0.066 & 0.132 & 0.197 & 0.262 & 0.326 \\
& \textbf{NELL-995-h100} &  0.195 &0.377 & 0.517 & 0.617 & 0.683 & 0.099 & 0.198 & 0.291 &  0.382 & 0.463 & 0.065 & 0.131 & 0.196 & 0.260 & 0.322 \\
\bottomrule
\end{tabular}
}
\label{table:others}
\end{table}

\begin{figure}[ht]
    \centering
    \includegraphics[clip,scale=0.28]{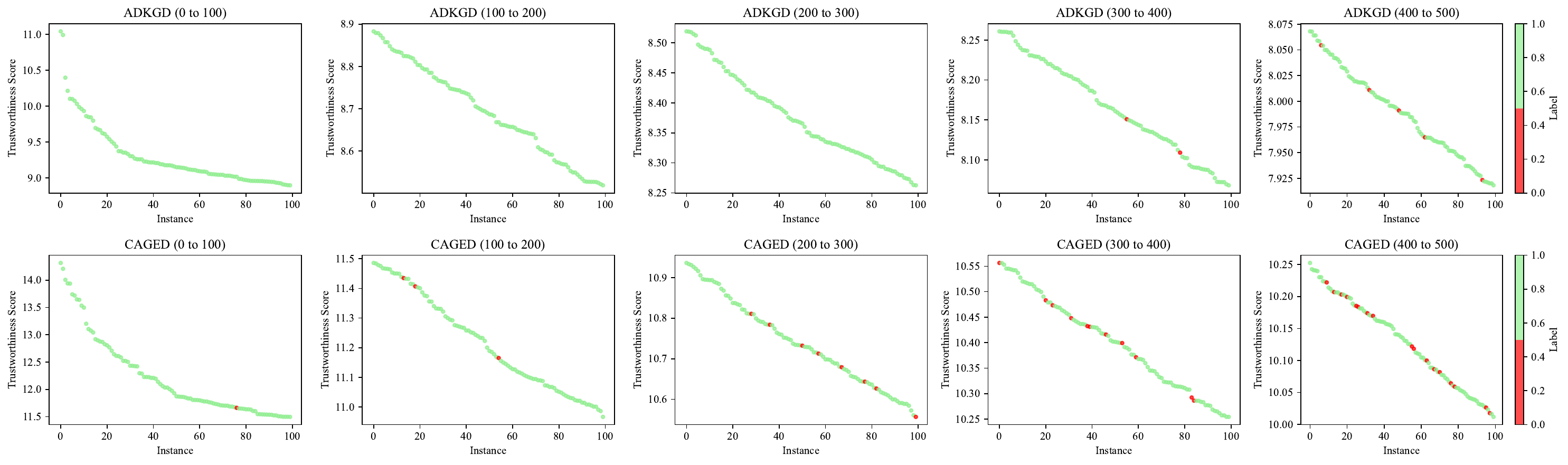}
    \caption{Analysis of Scores Among Nodes: ADKGD vs. CAGED on WN18RR.}
    \label{fig:distribution}
\end{figure}

Figure \ref{fig:distribution} compares the scores of nodes as detected by ADKGD and CAGED on the WN18RR dataset. The scores are plotted for five different intervals of 100 instances each, allowing for a detailed comparison of the methods across different segments of the dataset. This figure specifically contrasts the top 500 node scores. It is evident that in the top 200 nodes with the highest scores, the judgments made by ADKGD are entirely accurate, meaning that for the top 200 nodes, there are no misclassifications. In contrast, CAGED shows a few misclassifications, as indicated by the presence of red points. For the subsequent 300 nodes, ADKGD still demonstrates superior performance with fewer misclassifications compared to CAGED. This trend can be observed in the progressively lower scores and fewer errors as the nodes with lower scores are considered. In summary, ADKGD not only achieves perfect accuracy for the highest-scoring 200 nodes but also maintains a lower misclassification rate for the remaining 300 nodes, indicating its robustness and reliability in anomaly detection across different segments of the dataset.

\subsection{Ablation study (Q2)} 

The results presented in Table \ref{table:ablation_study} show the effectiveness of each component in the ADKGD on the WN18RR dataset. These results highlight the contributions of internal learning, neighbor aggregation, and the combination of views, as well as the impact of introducing KL divergence for consistency learning.
First, in the entity-view component, we observe that internal learning and neighbor aggregation independently contribute similarly to the anomaly detection performance. For example, removing internal learning results in average precision and recall across all $K$ values (1\%-5\%) of 0.605 and 0.331, respectively, while removing neighbor aggregation yields slightly similar average precision and recall of 0.606 and 0.332. These findings suggest that both internal learning and neighbor aggregation are critical and complementary in the entity-view's performance.
Integrating both internal learning and neighbor aggregation within the entity-view improves the average precision and recall to 0.635 and 0.345, respectively. This demonstrates that combining these two mechanisms enhances the model's capacity to effectively detect anomalies by leveraging both the internal structure and the surrounding neighborhood context.
When moving to the triplet-view, the average precision and recall improve further to 0.650 and 0.349, respectively, representing a 2.4\% and 1.2\% improvement over the combined entity-view alone. This indicates that analyzing relationships between triples provides additional insights critical for anomaly detection.
Combining the entity-view and triplet-view without KL divergence leads to further performance gains. Specifically, the average precision rises to 0.714, and the average recall increases to 0.370. However, introducing KL divergence for consistency learning between the two views achieves the best results, with an average precision of 0.741 and an average recall of 0.406. This represents a 14.0\% improvement in precision and a 16.3\% improvement in recall compared to the triplet-view component alone.
At $K$ = 1\%, the benefits of combining both views with KL divergence are even more apparent. The integrated model achieves the highest precision of 0.943 and recall of 0.188, outperforming all other configurations. This represents a 7.8\% improvement in precision and a 9.9\% improvement in recall over the triplet-view alone, highlighting the value of consistency learning.
In conclusion, while both internal learning and neighbor aggregation independently contribute to anomaly detection, their combination yields better results. Similarly, the integration of entity-view and triplet-view without KL divergence shows notable improvements, but the best performance is achieved when combining both views with internal learning, neighbor aggregation, and KL divergence for consistency learning. These findings emphasize the importance of leveraging multiple perspectives and integrating them cohesively for optimal anomaly detection in knowledge graphs.

\begin{table}[h!]
\centering
\caption{Effectiveness of each component on the WN18RR dataset}
\label{table:ablation_study}
\resizebox{\textwidth}{!}{%
\begin{tabular}{lcccccccccc}
\toprule
\multirow{3}{*}{Component} & \multicolumn{10}{c}{WN18RR} \\
\cmidrule(lr){2-11}
& \multicolumn{5}{c}{Precision@K} & \multicolumn{5}{c}{Recall@K} \\
\cmidrule(lr){2-6} \cmidrule(lr){7-11}
& 1\% & 2\% & 3\% & 4\% & 5\% & 1\% & 2\% & 3\% & 4\% & 5\% \\
\midrule
Entity-view w/o internal learning & 0.831 & 0.673 & 0.521 & 0.522 &0.471 & 0.153 & 0.273 & 0.354 & 0.411 & 0.466 \\
Entity-view w/o neighbors aggregation & 0.838 & 0.682 & 0.534 & 0.517 & 0.470 & 0.158 & 0.270 & 0.355 & 0.421 & 0.473 \\
Entity-view & 0.847 & 0.702 & 0.608 & 0.537 & 0.482 & 0.169 & 0.281 & 0.364 & 0.430 & 0.482 \\
Entity-view w/o internal learning & 0.838 & 0.705 & 0.611 & 0.532 & 0.473 & 0.163 & 0.271 & 0.357 & 0.422 & 0.477 \\
Entity-view w/o neighbors aggregation & 0.857 & 0.700 & 0.613 & 0.535 & 0.478 & 0.161 & 0.269 & 0.351 & 0.421 & 0.473 \\
triplet-view & 0.875 & 0.716 & 0.622 & 0.548 & 0.488 & 0.171 & 0.285 & 0.368 & 0.435 & 0.486 \\
Entity-view + triplet-view  w/o KL & 0.923 & 0.821 & 0.708 & 0.607 & 0.512 & 0.169 & 0.281 & 0.364 & 0.418 & 0.482 \\
Entity-view + triplet-view (KL) & 0.943 & 0.845 & 0.724 & 0.635 & 0.560 & 0.188 & 0.338 & 0.434 & 0.508 & 0.560 \\
\bottomrule
\end{tabular}
}
\end{table}

The results presented in Table \ref{table:aggregation_study} demonstrate that different aggregation methods for dual-view embeddings have a significant impact on the performance of anomaly detection in the WN18RR dataset. Among the tested methods—Hadamard product, summation, and concatenation followed by an MLP—concatenation with an MLP consistently outperforms the others in both Precision@K and Recall@K metrics across all evaluated $K$ values (1\%-5\%).
The Hadamard product, which combines embeddings through element-wise multiplication, shows the lowest performance in both precision and recall. While it effectively captures simple interactions between the embeddings, its inability to fully exploit the complementary information in the dual views results in suboptimal scores. For instance, at $K$ = 1\%, it achieves a Precision@K of 0.876 and a Recall@K of 0.172, which are notably lower than the other two methods.
Summation, a straightforward method of combining embeddings by directly adding them, improves upon the Hadamard product. It achieves a Precision@K of 0.890 and a Recall@K of 0.181 at $K$ = 1\%, with consistent improvements observed across all $K$ values. The performance gains suggest that summation is better at integrating the shared information between the two views compared to element-wise multiplication. However, it lacks the flexibility to model more complex relationships between the embeddings.
Concatenation followed by an MLP yields the best performance across all metrics. At $K$ = 1\%, it achieves a Precision@K of 0.943 and a Recall@K of 0.188, surpassing both the Hadamard product and summation by significant margins. The use of concatenation allows the embeddings from the two views to be fully preserved, and the MLP further optimizes the combined representation to maximize anomaly detection accuracy. This trainable approach provides greater expressiveness and adaptability, enabling the framework to capture nuanced patterns in the data.
Overall, the results indicate that concatenation with an MLP is the most effective aggregation method in the ADKGD. It significantly outperforms both summation and the Hadamard product, demonstrating that a learnable and flexible combination of dual-view embeddings is crucial for achieving superior anomaly detection performance.

\begin{table}[h!]
\centering
\caption{Effectiveness of different aggregation methods for dual-view embeddings on the WN18RR dataset}
\label{table:aggregation_study}
\resizebox{\textwidth}{!}{%
\begin{tabular}{lcccccccccc}
\toprule
\multirow{3}{*}{Aggregation Method} & \multicolumn{10}{c}{WN18RR} \\
\cmidrule(lr){2-11}
& \multicolumn{5}{c}{Precision@K} & \multicolumn{5}{c}{Recall@K} \\
\cmidrule(lr){2-6} \cmidrule(lr){7-11}
& 1\% & 2\% & 3\% & 4\% & 5\% & 1\% & 2\% & 3\% & 4\% & 5\% \\
\midrule
Hadamard product  & 0.876 & 0.782 & 0.668 & 0.578 & 0.489 & 0.172 & 0.286 & 0.369 & 0.437 & 0.487 \\
Summation & 0.890 & 0.805 & 0.684 & 0.593 & 0.501 & 0.181 & 0.298 & 0.374 & 0.443 & 0.492 \\
Concatenation + MLP & \textbf{0.943} & \textbf{0.845} & \textbf{0.724} & \textbf{0.635} & \textbf{0.560} & \textbf{0.188} & \textbf{0.338} & \textbf{0.434} & \textbf{0.508} & \textbf{0.560} \\
\bottomrule
\end{tabular}
}
\end{table}

\subsection{Hyper-parameter analysis (Q3)}

\begin{figure}[ht]
    \centering
    \includegraphics[clip,scale=0.30]{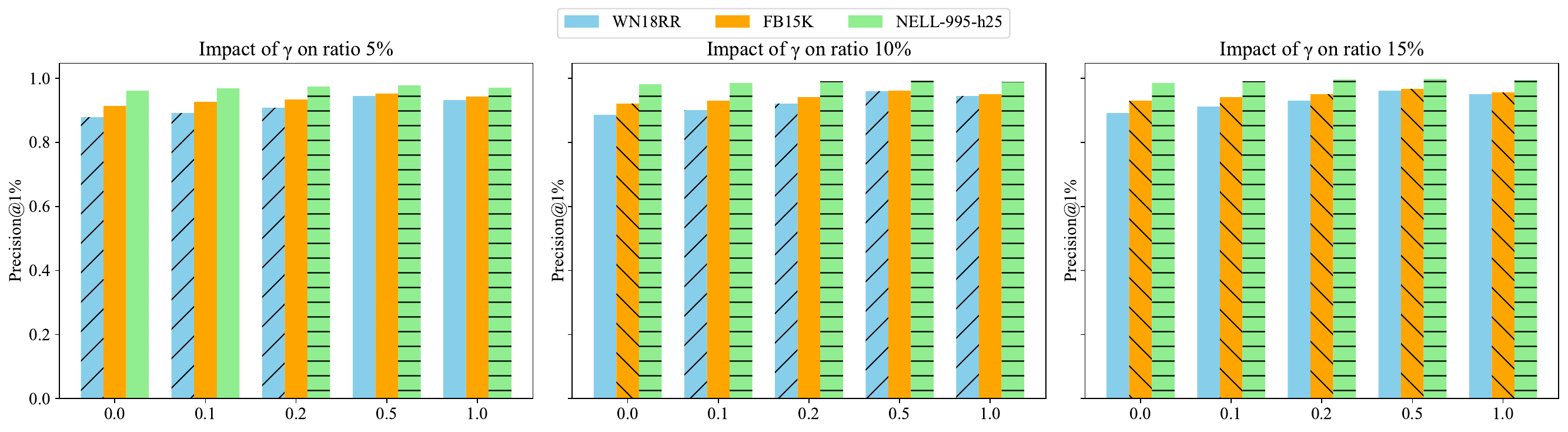}
    \caption{Impact of hyper-parameter 
    $\gamma$ on Precision@K when K = 1\% for FB15K, WN18RR, and NELL-995-h25 datasets at anomaly ratios of 5\%, 10\%, and 15\%.}
    \label{fig:margin}
\end{figure}

According to Figure \ref{fig:margin}, we can observe the overall impact of the hyper-parameter $ \gamma $ on Precision@1\% across three datasets (FB15K, WN18RR, and NELL-995-h25). Experiments are conducted under different anomaly ratios (5\%, 10\%, 15\%) to evaluate the effect of varying $ \gamma $ on model performance. Overall, as $ \gamma $ increases from 0.0 to 0.5, the average Precision@1\% across all datasets and anomaly ratios significantly improves. When $ \gamma $ increases from 0.0 to 0.5, the average Precision@1\% increases by approximately 10\%-15\%. However, when $ \gamma $ increases from 0.5 to 1.0, the Precision@1\% shows only a slight decrease, with a drop of around 1\%-3\%. Specifically, under a low anomaly ratio (5\%), Precision@1\% significantly improves as $ \gamma $ increases from 0.0 to 0.5, followed by a slight decrease at $ \gamma$ = 1.0. For a medium anomaly ratio (10\%), Precision@1\% reaches its peak at $ \gamma$ = 0.5. The improvement from $ \gamma$ = 0.0 to $ \gamma$ = 0.5 is substantial, followed by a slight decrease. Under a high anomaly ratio (15\%), the trend is similar to the low and medium anomaly ratios. Precision@1\% peaks at $ \gamma$ = 0.5, with a significant increase from $ \gamma$ = 0.0 to $ \gamma$ = 0.5, followed by a slight decrease at $ \gamma$ = 1.0. Experiments across different anomaly ratios indicate that $ \gamma$ = 0.5 is the optimal value for all datasets and anomaly ratios. This result highlights the importance of balancing the influence of the entity-view and triplet-view in the ADKGD method. Consistent data shows that adjusting $ \gamma $ to 0.5 yields the best results in terms of Precision@1\%, demonstrating the robustness and effectiveness of this hyper-parameter setting.

\begin{figure}[ht]
    \centering
    \includegraphics[clip,scale=0.30]{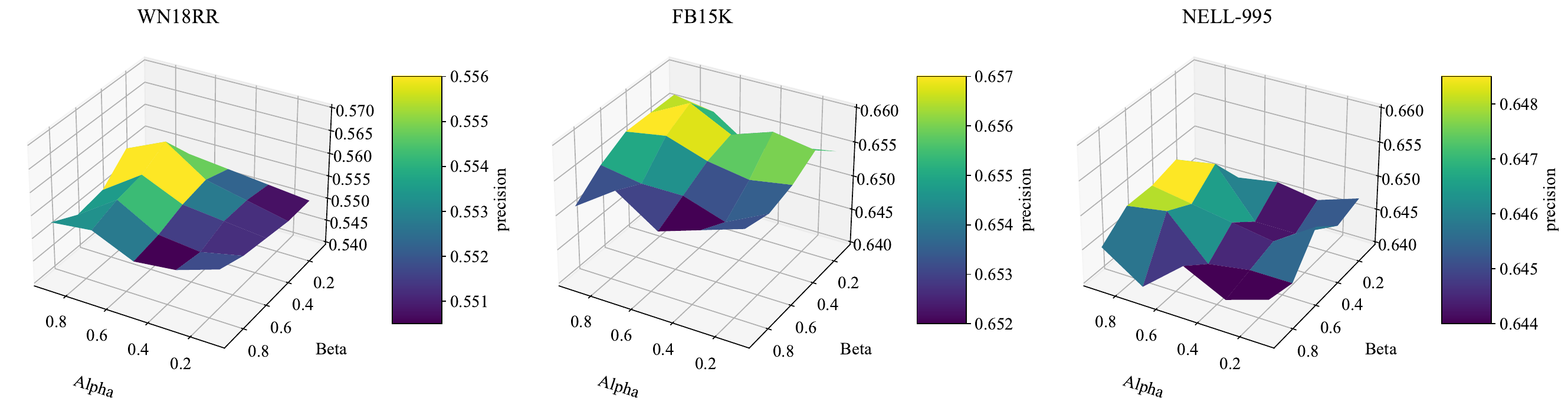}
    \caption{The hyperparameter tuning of $\alpha$ and $\beta$ on three datasets: WN18RR, FB15K, and NELL-995. The figures show the precision values for different combinations of $\alpha$ and $\beta$. The optimal values were $\alpha$ = 0.9 and $\beta$ = 0.3, resulting in the highest precision for each dataset.}
    \label{fig:3d}
\end{figure}

Figure \ref{fig:3d} illustrates the plots of precision values for various combinations of $\alpha$ and $\beta$ on the WN18RR, FB15K, and NELL-995 datasets. The $\alpha$ parameter controls the weight between internal information aggregation and aggregation with neighbor triplets, whereas the $\beta$ parameter balances the KL divergence loss and the dual-channel training. From these results, it is evident that the highest precision values for all three datasets are achieved when $\alpha$ = 0.9 and $\beta$ = 0.3. Specifically, the highest precision value for WN18RR is 0.560, for FB15K is 0.659, and for NELL-995 is 0.650. These results demonstrate the significance of the \textit{BI-LSTM} layer in capturing meaningful representations, as indicated by the high value of $\alpha$. Moreover, the considerable value of $\beta$ highlights the importance of the KL-loss in maintaining consistency between the entity-view and triplet-view representations. The optimal $\beta$ value of 0.3 suggests that a balanced contribution from both the primary loss and the consistency loss is crucial for achieving optimal performance. The findings imply that the cross-layer \textit{BI-LSTM} is particularly effective in distinguishing correct triplets from anomalies, and the KL-loss plays a vital role in aligning the representations from different perspectives.

\begin{figure}[ht]
    \centering
    \includegraphics[clip,scale=0.25]{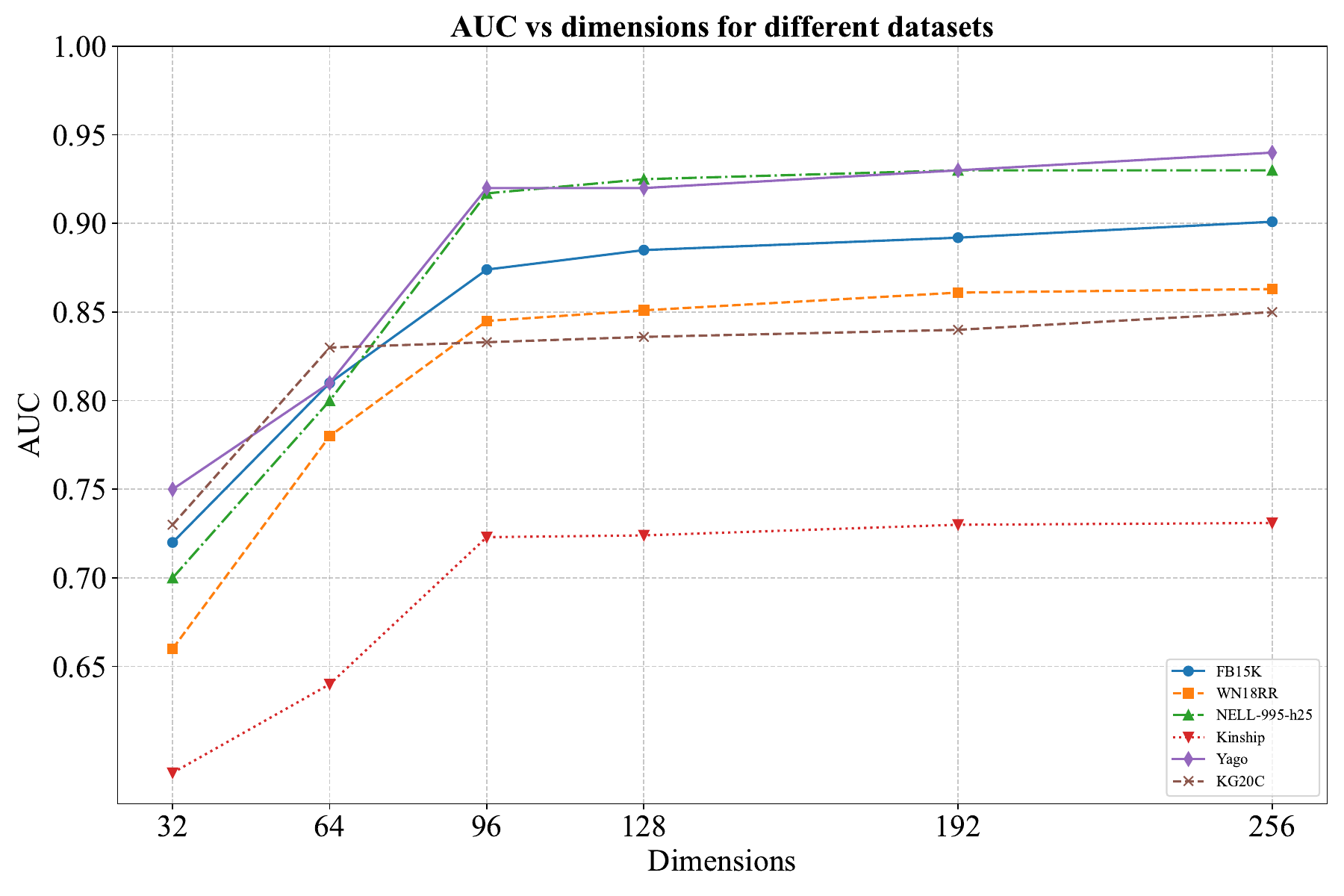}
    \caption{Relationship between embedding dimensions and AUC scores on six datasets. AUC scores peak between dimensions 96 and 128, with minimal growth beyond this range.}
    \label{fig:dimensions}
\end{figure}

The results presented in Figure \ref{fig:dimensions} illustrate the relationship between the embedding dimensions and the AUC scores on six datasets (FB15K, WN18RR, NELL-995-h25, Kinship, Yago, and KG20C). For all datasets, the AUC increases as the embedding dimension grows, reaching a peak in the range of dimensions between 96 and 128. Beyond this range, the AUC values exhibit minimal growth, suggesting that larger dimensions do not significantly improve anomaly detection performance. For example, on the FB15K dataset, the AUC reaches 0.885 at dimension 128 and shows only a marginal increase to 0.892 at dimension 192. Similarly, on the WN18RR dataset, the AUC grows from 0.845 at dimension 96 to 0.851 at dimension 128 and remains relatively stable thereafter. These patterns are consistent across all datasets, demonstrating that the optimal embedding dimension lies within the range of 96 to 128. Based on these observations, we set the embedding dimension to 100, balancing computational efficiency with model performance. This choice ensures that the model achieves near-optimal anomaly detection performance without incurring unnecessary computational overhead from excessively high embedding dimensions.

\subsection{Time efficiency (Q4)}

\begin{figure}[ht]
    \centering
    \includegraphics[clip,scale=0.30]{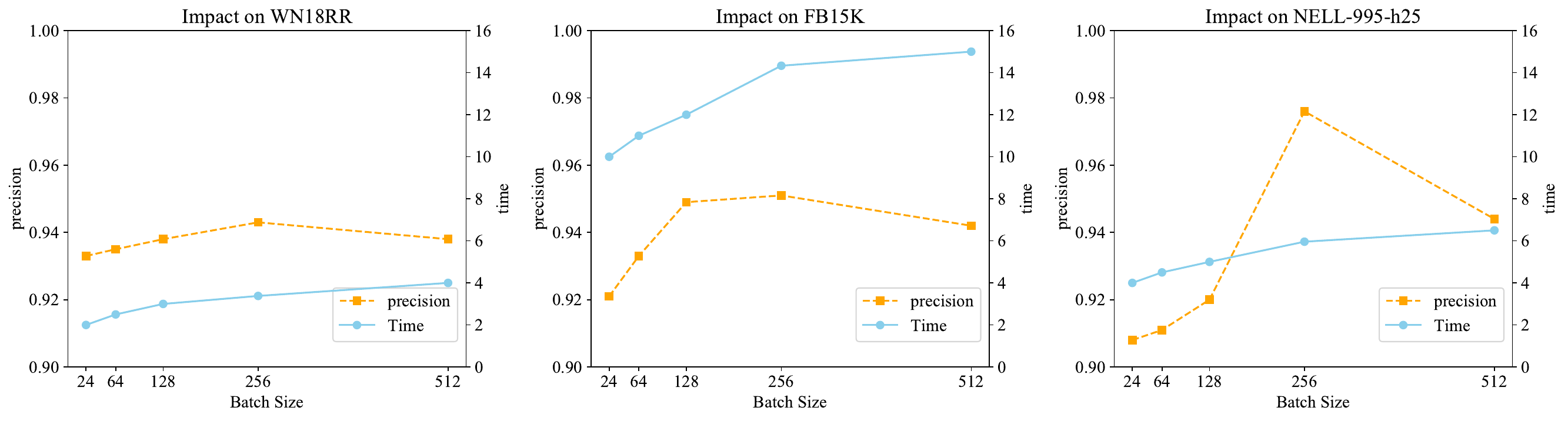}
    \caption{Impact of different batch sizes on precision and training time. }
    \label{fig:batch_size}
\end{figure}

Figure \ref{fig:batch_size} illustrates the impact of different batch sizes on precision and training time across three datasets: WN18RR, FB15K, and NELL-995-h25. Each figure represents one dataset, with the x-axis showing the batch size, the left y-axis showing the precision, and the right y-axis showing the training time in minutes. The data presented is based on the experiment with the 5\% anomaly ratio, specifically analyzing the precision results at K = 1\%.

For the WN18RR dataset, the precision increases with the batch size, peaking at 0.943 when the batch size is 256, before slightly decreasing at a batch size of 512. Correspondingly, the training time increases from about 2.0 minutes at a batch size of 24 to about 4.0 minutes at a batch size of 512. In the case of the FB15K dataset, the precision shows a significant rise from 0.921 at a batch size of 24, reaching a maximum of 0.951 at a batch size of 256, and then slightly decreasing at a batch size of 512. The training time follows a similar increasing trend, starting at 10.0 minutes and rising to 15.0 minutes as the batch size grows. For the NELL-995-h25 dataset, the precision starts at 0.908 for a batch size of 24 and peaks dramatically at 0.976 when the batch size is 256, but drops to 0.944 at a batch size of 512. The training time increases from 4.0 minutes to 6.5 minutes as the batch size increases.

From the above analysis, it is evident that a batch size of 256 provides the best balance between training time and precision performance for all three datasets. Smaller batch sizes result in shorter training times but lower accuracy, whereas larger batch sizes lead to longer training times and do not significantly improve or may even decrease accuracy. Thus, batch size 256 emerges as the optimal choice, ensuring high model accuracy while maintaining a reasonable training duration.

\begin{figure}[ht]
    \centering
    \includegraphics[clip,scale=0.30]{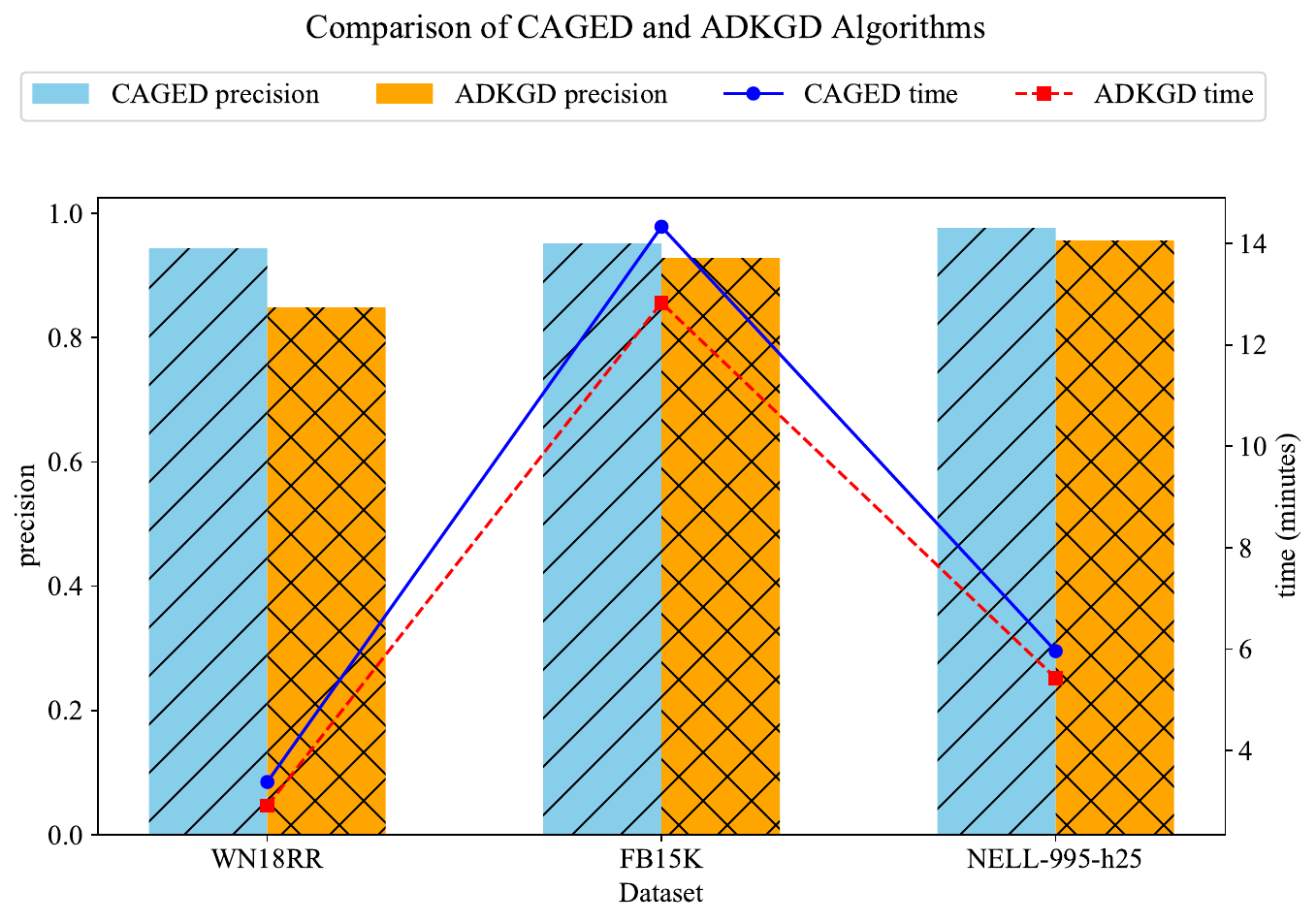}
    \caption{Comparison of CAGED and ADKGD in terms of time consumption and precision.}
    \label{fig:algorithms}
\end{figure}

We compared the performance of the CAGED and ADKGD algorithms in terms of time consumption and accuracy. The experimental results are summarized in Figure \ref{fig:algorithms}, which illustrates the performance on three datasets: WN18RR, FB15K, and NELL-995-H25. The time data represents the average time per epoch, while accuracy is measured by the precision value. The results indicate that the ADKGD algorithm consistently outperforms the CAGED algorithm in terms of accuracy across all three datasets. For instance, on the FB15K dataset, ADKGD achieves a precision of 0.951, compared to 0.927 for CAGED. This demonstrates the superior anomaly detection capabilities of the ADKGD algorithm. However, the enhanced accuracy of ADKGD comes at the cost of increased computational time. The time consumption for ADKGD is higher than that of the CAGED algorithm on all datasets. For example, on the FB15K dataset, the average epoch time for the ADKGD algorithm is 12.83 minutes, whereas the CAGED algorithm takes 14.33 minutes. While ADKGD requires more computational time, it significantly improves accuracy. However, it is an effective improvement over the CAGED algorithm. This trade-off between time consumption and accuracy suggests that ADKGD is particularly advantageous in scenarios where accuracy is of paramount importance. Therefore, the modifications introduced in ADKGD are validated as effective enhancements for anomaly detection tasks.

\section{Conclusion} \label{sec: conclusion}

Anomaly detection in KGs is crucial for maintaining the integrity and reliability of structured data used in various applications, such as recommendation systems, semantic search, and data integration. Traditional KG embedding methods like TransE, ComplEx, and DistMult have limitations in effectively identifying errors within KGs due to their inability to distinguish between normal and noisy triplets. To address these challenges, we propose a novel method called ADKGD. The ADKGD framework introduces a dual-channel training approach that integrates internal information aggregation and context information aggregation with entity-view and triplet-view. This method leverages \textit{BI-LSTM} networks to capture the internal relationships within triplets, and their neighboring context is determined by calculating the similarity of neighboring nodes. The key innovations of ADKGD include the use of two separate channels. One channel maintains the triplet-view, while the other reduces dimensionality to learn the entity-view of the data. Furthermore, a KL-based consistency loss is utilized to ensure coherence between the learned representations from both channels. The model constructs an anomaly scoring function that effectively differentiates between correct and anomaly triplets. By integrating dual-channel training and employing a consistency loss mechanism, ADKGD outperforms some traditional embedding-based methods and recent KG anomaly detection techniques. The robust performance of our model across different datasets and noise levels shows its effectiveness and reliability.

In the future, our research will focus on several aspects to further enhance the ADKGD framework. Exploring techniques to improve the scalability of ADKGD for very large-scale knowledge graphs is one potential direction \cite{ge2021largeea}. In addition, investigating real-time anomaly detection capabilities will allow ADKGD to be applied in dynamic and evolving data environments \cite{yan2021dynamic}. Furthermore, extending the application of ADKGD to integrate with large language models (LLMs) can significantly enhance the LLMs' capabilities. This combined approach ensures the LLM can generate higher quality responses in real-time \cite{agrawal2024cyberq}. Moreover, we can develop more efficient neighbor node search algorithms to further improve the framework’s computational efficiency and scalability. Combining this with advanced techniques such as diffusion models or generative adversarial networks (GAN) for learning and generating noise can also open new opportunities for improving anomaly detection and representation learning.

\section*{Acknowledgment}

This research was supported in part by the National Natural Science Foundation of China (No. 62272196), Guangzhou Basic and Applied Basic Research Foundation (No. 2024A04J9971), Engineering Research Center of Trustworthy AI, Ministry of Education (Jinan University), and Guangdong Key Laboratory of Data Security and Privacy Preserving.

\bibliographystyle{ACM-Reference-Format}
\bibliography{ADKGD.bib}


\begin{thebibliography}{48}


\ifx \showCODEN    \undefined \def \showCODEN     #1{\unskip}     \fi
\ifx \showDOI      \undefined \def \showDOI       #1{#1}\fi
\ifx \showISBNx    \undefined \def \showISBNx     #1{\unskip}     \fi
\ifx \showISBNxiii \undefined \def \showISBNxiii  #1{\unskip}     \fi
\ifx \showISSN     \undefined \def \showISSN      #1{\unskip}     \fi
\ifx \showLCCN     \undefined \def \showLCCN      #1{\unskip}     \fi
\ifx \shownote     \undefined \def \shownote      #1{#1}          \fi
\ifx \showarticletitle \undefined \def \showarticletitle #1{#1}   \fi
\ifx \showURL      \undefined \def \showURL       {\relax}        \fi
\providecommand\bibfield[2]{#2}
\providecommand\bibinfo[2]{#2}
\providecommand\natexlab[1]{#1}
\providecommand\showeprint[2][]{arXiv:#2}

\bibitem[Agrawal et~al\mbox{.}(2024)]%
        {agrawal2024cyberq}
\bibfield{author}{\bibinfo{person}{Garima Agrawal}, \bibinfo{person}{Kuntal Pal}, \bibinfo{person}{Yuli Deng}, \bibinfo{person}{Huan Liu}, {and} \bibinfo{person}{Ying-Chih Chen}.} \bibinfo{year}{2024}\natexlab{}.
\newblock \showarticletitle{{CyberQ}: Generating Questions and Answers for Cybersecurity Education Using Knowledge Graph-Augmented LLMs}. In \bibinfo{booktitle}{\emph{The AAAI Conference on Artificial Intelligence}}, Vol.~\bibinfo{volume}{38}. \bibinfo{pages}{23164--23172}.
\newblock


\bibitem[Balazevic et~al\mbox{.}(2019)]%
        {balazevic2019multi}
\bibfield{author}{\bibinfo{person}{Ivana Balazevic}, \bibinfo{person}{Carl Allen}, {and} \bibinfo{person}{Timothy Hospedales}.} \bibinfo{year}{2019}\natexlab{}.
\newblock \showarticletitle{Multi-relational poincar{\'e} graph embeddings}.
\newblock \bibinfo{journal}{\emph{Advances in Neural Information Processing Systems}}  \bibinfo{volume}{32} (\bibinfo{year}{2019}).
\newblock


\bibitem[Belth et~al\mbox{.}(2020)]%
        {belth2020normal}
\bibfield{author}{\bibinfo{person}{Caleb Belth}, \bibinfo{person}{Xinyi Zheng}, \bibinfo{person}{Jilles Vreeken}, {and} \bibinfo{person}{Danai Koutra}.} \bibinfo{year}{2020}\natexlab{}.
\newblock \showarticletitle{What is normal, what is strange, and what is missing in a knowledge graph: Unified characterization via inductive summarization}. In \bibinfo{booktitle}{\emph{The Web Conference}}. \bibinfo{pages}{1115--1126}.
\newblock


\bibitem[Bordes et~al\mbox{.}(2013)]%
        {bordes2013translating}
\bibfield{author}{\bibinfo{person}{Antoine Bordes}, \bibinfo{person}{Nicolas Usunier}, \bibinfo{person}{Alberto Garcia-Duran}, \bibinfo{person}{Jason Weston}, {and} \bibinfo{person}{Oksana Yakhnenko}.} \bibinfo{year}{2013}\natexlab{}.
\newblock \showarticletitle{Translating embeddings for modeling multi-relational data}.
\newblock \bibinfo{journal}{\emph{Advances in Neural Information Processing Systems}}  \bibinfo{volume}{26} (\bibinfo{year}{2013}).
\newblock


\bibitem[Cai et~al\mbox{.}(2021)]%
        {cai2021line}
\bibfield{author}{\bibinfo{person}{Lei Cai}, \bibinfo{person}{Jundong Li}, \bibinfo{person}{Jie Wang}, {and} \bibinfo{person}{Shuiwang Ji}.} \bibinfo{year}{2021}\natexlab{}.
\newblock \showarticletitle{Line graph neural networks for link prediction}.
\newblock \bibinfo{journal}{\emph{IEEE Transactions on Pattern Analysis and Machine Intelligence}} \bibinfo{volume}{44}, \bibinfo{number}{9} (\bibinfo{year}{2021}), \bibinfo{pages}{5103--5113}.
\newblock


\bibitem[Dettmers et~al\mbox{.}(2018)]%
        {dettmers2018convolutional}
\bibfield{author}{\bibinfo{person}{Tim Dettmers}, \bibinfo{person}{Pasquale Minervini}, \bibinfo{person}{Pontus Stenetorp}, {and} \bibinfo{person}{Sebastian Riedel}.} \bibinfo{year}{2018}\natexlab{}.
\newblock \showarticletitle{Convolutional {2D} knowledge graph embeddings}. In \bibinfo{booktitle}{\emph{The AAAI Conference on Artificial Intelligence}}, Vol.~\bibinfo{volume}{32}.
\newblock


\bibitem[Fei et~al\mbox{.}(2021)]%
        {fei2021enriching}
\bibfield{author}{\bibinfo{person}{Hao Fei}, \bibinfo{person}{Yafeng Ren}, \bibinfo{person}{Yue Zhang}, \bibinfo{person}{Donghong Ji}, {and} \bibinfo{person}{Xiaohui Liang}.} \bibinfo{year}{2021}\natexlab{}.
\newblock \showarticletitle{Enriching contextualized language model from knowledge graph for biomedical information extraction}.
\newblock \bibinfo{journal}{\emph{Briefings in Bioinformatics}} \bibinfo{volume}{22}, \bibinfo{number}{3} (\bibinfo{year}{2021}), \bibinfo{pages}{bbaa110}.
\newblock


\bibitem[Fei et~al\mbox{.}(2022)]%
        {fei2022matching}
\bibfield{author}{\bibinfo{person}{Hao Fei}, \bibinfo{person}{Shengqiong Wu}, \bibinfo{person}{Yafeng Ren}, {and} \bibinfo{person}{Meishan Zhang}.} \bibinfo{year}{2022}\natexlab{}.
\newblock \showarticletitle{Matching structure for dual learning}. In \bibinfo{booktitle}{\emph{International Conference on Machine Learning}}. PMLR, \bibinfo{pages}{6373--6391}.
\newblock


\bibitem[Ge et~al\mbox{.}(2021)]%
        {ge2021largeea}
\bibfield{author}{\bibinfo{person}{Congcong Ge}, \bibinfo{person}{Xiaoze Liu}, \bibinfo{person}{Lu Chen}, \bibinfo{person}{Yunjun Gao}, {and} \bibinfo{person}{Baihua Zheng}.} \bibinfo{year}{2021}\natexlab{}.
\newblock \showarticletitle{LargeEA: aligning entities for large-scale knowledge graphs}.
\newblock \bibinfo{journal}{\emph{The VLDB Endowment}} \bibinfo{volume}{15}, \bibinfo{number}{2} (\bibinfo{year}{2021}), \bibinfo{pages}{237--245}.
\newblock


\bibitem[Guo et~al\mbox{.}(2018)]%
        {guo2018knowledge}
\bibfield{author}{\bibinfo{person}{Shu Guo}, \bibinfo{person}{Quan Wang}, \bibinfo{person}{Lihong Wang}, \bibinfo{person}{Bin Wang}, {and} \bibinfo{person}{Li Guo}.} \bibinfo{year}{2018}\natexlab{}.
\newblock \showarticletitle{Knowledge graph embedding with iterative guidance from soft rules}. In \bibinfo{booktitle}{\emph{The AAAI Conference on Artificial Intelligence}}, Vol.~\bibinfo{volume}{32}.
\newblock


\bibitem[Hou et~al\mbox{.}(2017)]%
        {hou2017deep}
\bibfield{author}{\bibinfo{person}{Xianxu Hou}, \bibinfo{person}{Linlin Shen}, \bibinfo{person}{Ke Sun}, {and} \bibinfo{person}{Guoping Qiu}.} \bibinfo{year}{2017}\natexlab{}.
\newblock \showarticletitle{Deep feature consistent variational autoencoder}. In \bibinfo{booktitle}{\emph{IEEE Conference on Applications of Computer Vision}}. IEEE, \bibinfo{pages}{1133--1141}.
\newblock


\bibitem[Huang et~al\mbox{.}(2025)]%
        {huang2023survey}
\bibfield{author}{\bibinfo{person}{Lei Huang}, \bibinfo{person}{Weijiang Yu}, \bibinfo{person}{Weitao Ma}, \bibinfo{person}{Weihong Zhong}, \bibinfo{person}{Zhangyin Feng}, \bibinfo{person}{Haotian Wang}, \bibinfo{person}{Qianglong Chen}, \bibinfo{person}{Weihua Peng}, \bibinfo{person}{Xiaocheng Feng}, \bibinfo{person}{Bing Qin}, {et~al\mbox{.}}} \bibinfo{year}{2025}\natexlab{}.
\newblock \showarticletitle{A survey on hallucination in large language models: Principles, taxonomy, challenges, and open questions}.
\newblock \bibinfo{journal}{\emph{ACM Transactions on Information Systems}} (\bibinfo{year}{2025}), \bibinfo{pages}{1--54}.
\newblock


\bibitem[Huang et~al\mbox{.}(2019)]%
        {huang2019knowledge}
\bibfield{author}{\bibinfo{person}{Xiao Huang}, \bibinfo{person}{Jingyuan Zhang}, \bibinfo{person}{Dingcheng Li}, {and} \bibinfo{person}{Ping Li}.} \bibinfo{year}{2019}\natexlab{}.
\newblock \showarticletitle{Knowledge graph embedding based question answering}. In \bibinfo{booktitle}{\emph{The ACM International Conference on Web Search and Data Mining}}. \bibinfo{pages}{105--113}.
\newblock


\bibitem[Jia et~al\mbox{.}(2018)]%
        {jia2018pattern}
\bibfield{author}{\bibinfo{person}{Bin Jia}, \bibinfo{person}{Cailing Dong}, \bibinfo{person}{Zhijiang Chen}, \bibinfo{person}{Kuo-Chu Chang}, \bibinfo{person}{Nichole Sullivan}, {and} \bibinfo{person}{Genshe Chen}.} \bibinfo{year}{2018}\natexlab{}.
\newblock \showarticletitle{Pattern discovery and anomaly detection via knowledge graph}. In \bibinfo{booktitle}{\emph{The International Conference on Information Fusion}}. IEEE, \bibinfo{pages}{2392--2399}.
\newblock


\bibitem[Jia et~al\mbox{.}(2019)]%
        {jia2019triple}
\bibfield{author}{\bibinfo{person}{Shengbin Jia}, \bibinfo{person}{Yang Xiang}, \bibinfo{person}{Xiaojun Chen}, {and} \bibinfo{person}{Kun Wang}.} \bibinfo{year}{2019}\natexlab{}.
\newblock \showarticletitle{Triple trustworthiness measurement for knowledge graph}. In \bibinfo{booktitle}{\emph{The World Wide Web Conference}}. \bibinfo{pages}{2865--2871}.
\newblock


\bibitem[Lewis et~al\mbox{.}(2020)]%
        {lewis2020retrieval}
\bibfield{author}{\bibinfo{person}{Patrick Lewis}, \bibinfo{person}{Ethan Perez}, \bibinfo{person}{Aleksandra Piktus}, \bibinfo{person}{Fabio Petroni}, \bibinfo{person}{Vladimir Karpukhin}, \bibinfo{person}{Naman Goyal}, \bibinfo{person}{Heinrich K{\"u}ttler}, \bibinfo{person}{Mike Lewis}, \bibinfo{person}{Wen-tau Yih}, \bibinfo{person}{Tim Rockt{\"a}schel}, {et~al\mbox{.}}} \bibinfo{year}{2020}\natexlab{}.
\newblock \showarticletitle{Retrieval-augmented generation for knowledge-intensive nlp tasks}.
\newblock \bibinfo{journal}{\emph{Advances in Neural Information Processing Systems}}  \bibinfo{volume}{33} (\bibinfo{year}{2020}), \bibinfo{pages}{9459--9474}.
\newblock


\bibitem[Li et~al\mbox{.}(2024)]%
        {li2024ripple}
\bibfield{author}{\bibinfo{person}{Chen Li}, \bibinfo{person}{Yang Cao}, \bibinfo{person}{Ye Zhu}, \bibinfo{person}{Debo Cheng}, \bibinfo{person}{Chengyuan Li}, {and} \bibinfo{person}{Yasuhiko Morimoto}.} \bibinfo{year}{2024}\natexlab{}.
\newblock \showarticletitle{Ripple knowledge graph convolutional networks for recommendation systems}.
\newblock \bibinfo{journal}{\emph{Machine Intelligence Research}} (\bibinfo{year}{2024}), \bibinfo{pages}{1--14}.
\newblock


\bibitem[Li et~al\mbox{.}(2023)]%
        {li2023graph}
\bibfield{author}{\bibinfo{person}{Shiyang Li}, \bibinfo{person}{Yifan Gao}, \bibinfo{person}{Haoming Jiang}, \bibinfo{person}{Qingyu Yin}, \bibinfo{person}{Zheng Li}, \bibinfo{person}{Xifeng Yan}, \bibinfo{person}{Chao Zhang}, {and} \bibinfo{person}{Bing Yin}.} \bibinfo{year}{2023}\natexlab{}.
\newblock \showarticletitle{Graph Reasoning for Question Answering with Triplet Retrieval}. In \bibinfo{booktitle}{\emph{Findings of the Association for Computational Linguistics}}. \bibinfo{pages}{3366--3375}.
\newblock


\bibitem[Lin et~al\mbox{.}(2019)]%
        {lin2019kagnet}
\bibfield{author}{\bibinfo{person}{Bill~Yuchen Lin}, \bibinfo{person}{Xinyue Chen}, \bibinfo{person}{Jamin Chen}, {and} \bibinfo{person}{Xiang Ren}.} \bibinfo{year}{2019}\natexlab{}.
\newblock \showarticletitle{{KagNet}: Knowledge-Aware Graph Networks for Commonsense Reasoning}. In \bibinfo{booktitle}{\emph{The Conference on Empirical Methods in Natural Language Processing and the International Joint Conference on Natural Language Processing}}. \bibinfo{pages}{2829--2839}.
\newblock


\bibitem[Lin et~al\mbox{.}(2015)]%
        {lin2015learning}
\bibfield{author}{\bibinfo{person}{Yankai Lin}, \bibinfo{person}{Zhiyuan Liu}, \bibinfo{person}{Maosong Sun}, \bibinfo{person}{Yang Liu}, {and} \bibinfo{person}{Xuan Zhu}.} \bibinfo{year}{2015}\natexlab{}.
\newblock \showarticletitle{Learning entity and relation embeddings for knowledge graph completion}. In \bibinfo{booktitle}{\emph{The AAAI Conference on Artificial Intelligence}}, Vol.~\bibinfo{volume}{29}.
\newblock


\bibitem[Liu et~al\mbox{.}(2024a)]%
        {liu2024knowledge}
\bibfield{author}{\bibinfo{person}{Xiangyu Liu}, \bibinfo{person}{Yang Liu}, {and} \bibinfo{person}{Wei Hu}.} \bibinfo{year}{2024}\natexlab{a}.
\newblock \showarticletitle{Knowledge Graph Error Detection with Contrastive Confidence Adaption}. In \bibinfo{booktitle}{\emph{The AAAI Conference on Artificial Intelligence}}, Vol.~\bibinfo{volume}{38}. \bibinfo{pages}{8824--8831}.
\newblock


\bibitem[Liu et~al\mbox{.}(2024b)]%
        {liu2024sesicl}
\bibfield{author}{\bibinfo{person}{Xingyu Liu}, \bibinfo{person}{Jielong Tang}, \bibinfo{person}{Mengyang Li}, \bibinfo{person}{Junmei Han}, \bibinfo{person}{Gang Xiao}, {and} \bibinfo{person}{Jianchun Jiang}.} \bibinfo{year}{2024}\natexlab{b}.
\newblock \showarticletitle{{SeSICL}: Semantic and Structural Integrated Contrastive Learning for Knowledge Graph Error Detection}.
\newblock \bibinfo{journal}{\emph{IEEE Access}} (\bibinfo{year}{2024}).
\newblock


\bibitem[Liu et~al\mbox{.}(2023)]%
        {liu2023learning}
\bibfield{author}{\bibinfo{person}{Yixin Liu}, \bibinfo{person}{Kaize Ding}, \bibinfo{person}{Jianling Wang}, \bibinfo{person}{Vincent Lee}, \bibinfo{person}{Huan Liu}, {and} \bibinfo{person}{Shirui Pan}.} \bibinfo{year}{2023}\natexlab{}.
\newblock \showarticletitle{Learning strong graph neural networks with weak information}. In \bibinfo{booktitle}{\emph{The ACM SIGKDD Conference on Knowledge Discovery and Data Mining}}. \bibinfo{pages}{1559--1571}.
\newblock


\bibitem[Luo et~al\mbox{.}(2023)]%
        {luo2023dual}
\bibfield{author}{\bibinfo{person}{Zhenfei Luo}, \bibinfo{person}{Yixiang Dong}, \bibinfo{person}{Qinghua Zheng}, \bibinfo{person}{Huan Liu}, {and} \bibinfo{person}{Minnan Luo}.} \bibinfo{year}{2023}\natexlab{}.
\newblock \showarticletitle{Dual-channel graph contrastive learning for self-supervised graph-level representation learning}.
\newblock \bibinfo{journal}{\emph{Pattern Recognition}}  \bibinfo{volume}{139} (\bibinfo{year}{2023}), \bibinfo{pages}{109448}.
\newblock


\bibitem[Melo and Paulheim(2017)]%
        {melo2017detection}
\bibfield{author}{\bibinfo{person}{Andr{\'e} Melo} {and} \bibinfo{person}{Heiko Paulheim}.} \bibinfo{year}{2017}\natexlab{}.
\newblock \showarticletitle{Detection of relation assertion errors in knowledge graphs}. In \bibinfo{booktitle}{\emph{The 9th Knowledge Capture Conference}}. \bibinfo{pages}{1--8}.
\newblock


\bibitem[Ouyang et~al\mbox{.}(2022)]%
        {ouyang2022training}
\bibfield{author}{\bibinfo{person}{Long Ouyang}, \bibinfo{person}{Jeffrey Wu}, \bibinfo{person}{Xu Jiang}, \bibinfo{person}{Diogo Almeida}, \bibinfo{person}{Carroll Wainwright}, \bibinfo{person}{Pamela Mishkin}, \bibinfo{person}{Chong Zhang}, \bibinfo{person}{Sandhini Agarwal}, \bibinfo{person}{Katarina Slama}, \bibinfo{person}{Alex Ray}, {et~al\mbox{.}}} \bibinfo{year}{2022}\natexlab{}.
\newblock \showarticletitle{Training language models to follow instructions with human feedback}.
\newblock \bibinfo{journal}{\emph{Advances in Neural Information Processing Systems}}  \bibinfo{volume}{35} (\bibinfo{year}{2022}), \bibinfo{pages}{27730--27744}.
\newblock


\bibitem[Pellissier~Tanon et~al\mbox{.}(2017)]%
        {pellissier2017completeness}
\bibfield{author}{\bibinfo{person}{Thomas Pellissier~Tanon}, \bibinfo{person}{Daria Stepanova}, \bibinfo{person}{Simon Razniewski}, \bibinfo{person}{Paramita Mirza}, {and} \bibinfo{person}{Gerhard Weikum}.} \bibinfo{year}{2017}\natexlab{}.
\newblock \showarticletitle{Completeness-aware rule learning from knowledge graphs}. In \bibinfo{booktitle}{\emph{The International Semantic Web Conference}}. Springer, \bibinfo{pages}{507--525}.
\newblock


\bibitem[Petroni et~al\mbox{.}(2019)]%
        {petroni2019language}
\bibfield{author}{\bibinfo{person}{Fabio Petroni}, \bibinfo{person}{Tim Rockt{\"a}schel}, \bibinfo{person}{Sebastian Riedel}, \bibinfo{person}{Patrick Lewis}, \bibinfo{person}{Anton Bakhtin}, \bibinfo{person}{Yuxiang Wu}, {and} \bibinfo{person}{Alexander Miller}.} \bibinfo{year}{2019}\natexlab{}.
\newblock \showarticletitle{Language Models as Knowledge Bases?}. In \bibinfo{booktitle}{\emph{The Conference on Empirical Methods in Natural Language Processing and the 9th International Joint Conference on Natural Language Processing}}. \bibinfo{pages}{2463--2473}.
\newblock


\bibitem[Robinson et~al\mbox{.}(2016)]%
        {robinson2016families}
\bibfield{author}{\bibinfo{person}{Joseph~P Robinson}, \bibinfo{person}{Ming Shao}, \bibinfo{person}{Yue Wu}, {and} \bibinfo{person}{Yun Fu}.} \bibinfo{year}{2016}\natexlab{}.
\newblock \showarticletitle{Families in the wild ({FIW}) large-scale kinship image database and benchmarks}. In \bibinfo{booktitle}{\emph{The ACM International Conference on Multimedia}}. \bibinfo{pages}{242--246}.
\newblock


\bibitem[Saxena et~al\mbox{.}(2020)]%
        {saxena2020improving}
\bibfield{author}{\bibinfo{person}{Apoorv Saxena}, \bibinfo{person}{Aditay Tripathi}, {and} \bibinfo{person}{Partha Talukdar}.} \bibinfo{year}{2020}\natexlab{}.
\newblock \showarticletitle{Improving multi-hop question answering over knowledge graphs using knowledge base embeddings}. In \bibinfo{booktitle}{\emph{The Annual Meeting of the Association for Computational Linguistics}}. \bibinfo{pages}{4498--4507}.
\newblock


\bibitem[Shan et~al\mbox{.}(2018)]%
        {shan2018confidence}
\bibfield{author}{\bibinfo{person}{Yingchun Shan}, \bibinfo{person}{Chenyang Bu}, \bibinfo{person}{Xiaojian Liu}, \bibinfo{person}{Shengwei Ji}, {and} \bibinfo{person}{Lei Li}.} \bibinfo{year}{2018}\natexlab{}.
\newblock \showarticletitle{Confidence-aware negative sampling method for noisy knowledge graph embedding}. In \bibinfo{booktitle}{\emph{IEEE International Conference on Big Knowledge}}. IEEE, \bibinfo{pages}{33--40}.
\newblock


\bibitem[Sheth et~al\mbox{.}(2019)]%
        {sheth2019knowledge}
\bibfield{author}{\bibinfo{person}{Amit Sheth}, \bibinfo{person}{Swati Padhee}, {and} \bibinfo{person}{Amelie Gyrard}.} \bibinfo{year}{2019}\natexlab{}.
\newblock \showarticletitle{Knowledge graphs and knowledge networks: the story in brief}.
\newblock \bibinfo{journal}{\emph{IEEE Internet Computing}} \bibinfo{volume}{23}, \bibinfo{number}{4} (\bibinfo{year}{2019}), \bibinfo{pages}{67--75}.
\newblock


\bibitem[Suchanek et~al\mbox{.}(2007)]%
        {yago}
\bibfield{author}{\bibinfo{person}{Fabian~M Suchanek}, \bibinfo{person}{Gjergji Kasneci}, {and} \bibinfo{person}{Gerhard Weikum}.} \bibinfo{year}{2007}\natexlab{}.
\newblock \showarticletitle{{YAGO}: a core of semantic knowledge}. In \bibinfo{booktitle}{\emph{The 16th International Conference on World Wide Web}}. \bibinfo{pages}{697--706}.
\newblock


\bibitem[Toutanova et~al\mbox{.}(2015)]%
        {toutanova2015representing}
\bibfield{author}{\bibinfo{person}{Kristina Toutanova}, \bibinfo{person}{Danqi Chen}, \bibinfo{person}{Patrick Pantel}, \bibinfo{person}{Hoifung Poon}, \bibinfo{person}{Pallavi Choudhury}, {and} \bibinfo{person}{Michael Gamon}.} \bibinfo{year}{2015}\natexlab{}.
\newblock \showarticletitle{Representing text for joint embedding of text and knowledge bases}. In \bibinfo{booktitle}{\emph{The Conference on Empirical Methods in Natural Language Processing}}. \bibinfo{pages}{1499--1509}.
\newblock


\bibitem[Tran and Takasu(2019)]%
        {tran_exploringscholarlydata_2019}
\bibfield{author}{\bibinfo{person}{Hung~Nghiep Tran} {and} \bibinfo{person}{Atsuhiro Takasu}.} \bibinfo{year}{2019}\natexlab{}.
\newblock \showarticletitle{Exploring scholarly data by semantic query on knowledge graph embedding space}. In \bibinfo{booktitle}{\emph{The 23rd International Conference on Theory and Practice of Digital Libraries}}. Springer, \bibinfo{pages}{154--162}.
\newblock


\bibitem[Trouillon et~al\mbox{.}(2016)]%
        {trouillon2016complex}
\bibfield{author}{\bibinfo{person}{Th{\'e}o Trouillon}, \bibinfo{person}{Johannes Welbl}, \bibinfo{person}{Sebastian Riedel}, \bibinfo{person}{{\'E}ric Gaussier}, {and} \bibinfo{person}{Guillaume Bouchard}.} \bibinfo{year}{2016}\natexlab{}.
\newblock \showarticletitle{Complex embeddings for simple link prediction}. In \bibinfo{booktitle}{\emph{International Conference on Machine Learning}}. PMLR, \bibinfo{pages}{2071--2080}.
\newblock


\bibitem[Wang et~al\mbox{.}(2023)]%
        {wang2023boosting}
\bibfield{author}{\bibinfo{person}{Jianing Wang}, \bibinfo{person}{Qiushi Sun}, \bibinfo{person}{Nuo Chen}, \bibinfo{person}{Xiang Li}, {and} \bibinfo{person}{Ming Gao}.} \bibinfo{year}{2023}\natexlab{}.
\newblock \showarticletitle{Boosting language models reasoning with chain-of-knowledge prompting}.
\newblock \bibinfo{journal}{\emph{arXiv preprint arXiv:2306.06427}} (\bibinfo{year}{2023}).
\newblock


\bibitem[Wu et~al\mbox{.}(2023a)]%
        {wu2023ai}
\bibfield{author}{\bibinfo{person}{Jiayang Wu}, \bibinfo{person}{Wensheng Gan}, \bibinfo{person}{Zefeng Chen}, \bibinfo{person}{Shicheng Wan}, {and} \bibinfo{person}{Hong Lin}.} \bibinfo{year}{2023}\natexlab{a}.
\newblock \showarticletitle{Ai-generated content ({AIGC}): A survey}.
\newblock \bibinfo{journal}{\emph{arXiv preprint arXiv:2304.06632}} (\bibinfo{year}{2023}).
\newblock


\bibitem[Wu et~al\mbox{.}(2023b)]%
        {wu2023multimodal}
\bibfield{author}{\bibinfo{person}{Jiayang Wu}, \bibinfo{person}{Wensheng Gan}, \bibinfo{person}{Zefeng Chen}, \bibinfo{person}{Shicheng Wan}, {and} \bibinfo{person}{Philip~S Yu}.} \bibinfo{year}{2023}\natexlab{b}.
\newblock \showarticletitle{Multimodal large language models: A survey}. In \bibinfo{booktitle}{\emph{IEEE International Conference on Big Data}}. IEEE, \bibinfo{pages}{2247--2256}.
\newblock


\bibitem[Wu et~al\mbox{.}(2020)]%
        {wu2020comprehensive}
\bibfield{author}{\bibinfo{person}{Zonghan Wu}, \bibinfo{person}{Shirui Pan}, \bibinfo{person}{Fengwen Chen}, \bibinfo{person}{Guodong Long}, \bibinfo{person}{Chengqi Zhang}, {and} \bibinfo{person}{Philip~S Yu}.} \bibinfo{year}{2020}\natexlab{}.
\newblock \showarticletitle{A comprehensive survey on graph neural networks}.
\newblock \bibinfo{journal}{\emph{IEEE Transactions on Neural Networks and Learning Systems}} \bibinfo{volume}{32}, \bibinfo{number}{1} (\bibinfo{year}{2020}), \bibinfo{pages}{4--24}.
\newblock


\bibitem[Xu et~al\mbox{.}(2024)]%
        {xu2024hallucination}
\bibfield{author}{\bibinfo{person}{Ziwei Xu}, \bibinfo{person}{Sanjay Jain}, {and} \bibinfo{person}{Mohan Kankanhalli}.} \bibinfo{year}{2024}\natexlab{}.
\newblock \showarticletitle{Hallucination is inevitable: An innate limitation of large language models}.
\newblock \bibinfo{journal}{\emph{arXiv preprint arXiv:2401.11817}} (\bibinfo{year}{2024}).
\newblock


\bibitem[Yan et~al\mbox{.}(2021)]%
        {yan2021dynamic}
\bibfield{author}{\bibinfo{person}{Yuchen Yan}, \bibinfo{person}{Lihui Liu}, \bibinfo{person}{Yikun Ban}, \bibinfo{person}{Baoyu Jing}, {and} \bibinfo{person}{Hanghang Tong}.} \bibinfo{year}{2021}\natexlab{}.
\newblock \showarticletitle{Dynamic knowledge graph alignment}. In \bibinfo{booktitle}{\emph{The AAAI Conference on Artificial Intelligence}}, Vol.~\bibinfo{volume}{35}. \bibinfo{pages}{4564--4572}.
\newblock


\bibitem[Yang et~al\mbox{.}(2015)]%
        {yang2015embedding}
\bibfield{author}{\bibinfo{person}{Bishan Yang}, \bibinfo{person}{Scott Wen-tau Yih}, \bibinfo{person}{Xiaodong He}, \bibinfo{person}{Jianfeng Gao}, {and} \bibinfo{person}{Li Deng}.} \bibinfo{year}{2015}\natexlab{}.
\newblock \showarticletitle{Embedding Entities and Relations for Learning and Inference in Knowledge Bases}. In \bibinfo{booktitle}{\emph{the International Conference on Learning Representations}}.
\newblock


\bibitem[Yao et~al\mbox{.}(2023)]%
        {yao2023viskop}
\bibfield{author}{\bibinfo{person}{Zijun Yao}, \bibinfo{person}{Yuanyong Chen}, \bibinfo{person}{Xin Lv}, \bibinfo{person}{Shulin Cao}, \bibinfo{person}{Amy Xin}, \bibinfo{person}{Jifan Yu}, \bibinfo{person}{Hailong Jin}, \bibinfo{person}{Jianjun Xu}, \bibinfo{person}{Peng Zhang}, \bibinfo{person}{Lei Hou}, {et~al\mbox{.}}} \bibinfo{year}{2023}\natexlab{}.
\newblock \showarticletitle{{VisKoP}: Visual Knowledge oriented Programming for Interactive Knowledge Base Question Answering}. In \bibinfo{booktitle}{\emph{The Annual Meeting of the Association for Computational Linguistic}}. \bibinfo{pages}{179--189}.
\newblock


\bibitem[Zhang et~al\mbox{.}(2022)]%
        {zhang2022contrastive}
\bibfield{author}{\bibinfo{person}{Qinggang Zhang}, \bibinfo{person}{Junnan Dong}, \bibinfo{person}{Keyu Duan}, \bibinfo{person}{Xiao Huang}, \bibinfo{person}{Yezi Liu}, {and} \bibinfo{person}{Linchuan Xu}.} \bibinfo{year}{2022}\natexlab{}.
\newblock \showarticletitle{Contrastive knowledge graph error detection}. In \bibinfo{booktitle}{\emph{The 31st ACM International Conference on Information \& Knowledge Management}}. \bibinfo{pages}{2590--2599}.
\newblock


\bibitem[Zhang et~al\mbox{.}(2023)]%
        {zhang2023line}
\bibfield{author}{\bibinfo{person}{Zehua Zhang}, \bibinfo{person}{Shilin Sun}, \bibinfo{person}{Guixiang Ma}, {and} \bibinfo{person}{Caiming Zhong}.} \bibinfo{year}{2023}\natexlab{}.
\newblock \showarticletitle{Line graph contrastive learning for link prediction}.
\newblock \bibinfo{journal}{\emph{Pattern Recognition}}  \bibinfo{volume}{140} (\bibinfo{year}{2023}), \bibinfo{pages}{109537}.
\newblock


\bibitem[Zheng et~al\mbox{.}(2020)]%
        {zheng2020dgtn}
\bibfield{author}{\bibinfo{person}{Yujia Zheng}, \bibinfo{person}{Siyi Liu}, \bibinfo{person}{Zekun Li}, {and} \bibinfo{person}{Shu Wu}.} \bibinfo{year}{2020}\natexlab{}.
\newblock \showarticletitle{{DGTN}: Dual-channel graph transition network for session-based recommendation}. In \bibinfo{booktitle}{\emph{International Conference on Data Mining Workshops}}. IEEE, \bibinfo{pages}{236--242}.
\newblock


\bibitem[Zhou et~al\mbox{.}(2016)]%
        {zhou2016attention}
\bibfield{author}{\bibinfo{person}{Peng Zhou}, \bibinfo{person}{Wei Shi}, \bibinfo{person}{Jun Tian}, \bibinfo{person}{Zhenyu Qi}, \bibinfo{person}{Bingchen Li}, \bibinfo{person}{Hongwei Hao}, {and} \bibinfo{person}{Bo Xu}.} \bibinfo{year}{2016}\natexlab{}.
\newblock \showarticletitle{Attention-based bidirectional long short-term memory networks for relation classification}. In \bibinfo{booktitle}{\emph{The Association for Computational Linguistics}}. \bibinfo{pages}{207--212}.
\newblock


\end{thebibliography}

\clearpage
\section*{Appendix}

Table \ref{tab1} illustrates the predicted anomaly scores computed by our algorithm and the corresponding true labels for the triples on the dataset FB15K. The labels range from 1 for anomalous triples to 0 for non-anomalous triples. For the first 10 triples ($T_1$ to $T_{10}$), labeled as anomalies (label = 1), the predicted scores are notably higher, indicating a strong confidence in their anomaly status. On the other hand, for triples from $T_{93,025}$ to $T_{93,034}$, labeled as non-anomalous (label = 0), the predicted scores are significantly lower. This stark contrast in scores demonstrates the effectiveness of our algorithm in assigning appropriate scores that reflect the anomaly status of triples. Specifically, the average predicted score for the first 10 anomalous triples ($T_1$ to $T_{10}$) is 9.6464, while the average predicted score for the non-anomalous triples ($T_{93,025}$ to $T_{93,034}$) is 5.6767. The difference in the average predicted score between these two groups is 3.9697. This substantial gap further validates our algorithm's ability to accurately score triples and distinguish anomalies from non-anomalies based on their inherent features. The triple descriptions provided below correspond to the real triples from the datasets. These descriptions illustrate the specific entities, relations, and associated details for each triple, allowing us to better understand the context of the anomalous and non-anomalous triples identified by the model. For example, $T_1$ represents a triple related to GDP nominal per capita in a specific statistical region, while $T_{93,025}$ corresponds to a film distributor relationship. The contextual differences between these triples help explain why the model assigns higher scores to the former, given its anomalous nature, and lower scores to the latter, reflecting its regular pattern within the graph. By leveraging the unique structural patterns of the knowledge graph, our model successfully identifies triples with anomalous behaviors and assigns them higher scores, demonstrating its robustness and reliability in anomaly detection tasks.

\appendix

\begin{table}[ht]
\centering
\small
\caption{Predicted anomaly scores and true labels for triples on the dataset FB15K. The first 10 triples ($T_1$ to $T_{10}$) are labeled as anomalous (label = 1) and exhibit higher predicted scores, while the last 10 triples ($T_{93,025}$ to $T_{93,034}$) are labeled as non-anomalous (label = 0) and have significantly lower scores. This demonstrates the model's ability to distinguish anomalies based on their scores.}
\begin{tabular}{l c c l c c}
\toprule
\textbf{Triple ID} & \textbf{Predicted Score} & \textbf{True Label} & \textbf{Triple ID} & \textbf{Predicted Score} & \textbf{True Label} \\ 
\midrule
$T_1$   & 10.2339 & 1 & $T_{93,025}$  & 5.6768 & 0 \\ 
$T_2$   & 10.0615 & 1 & $T_{93,026}$  & 5.6768 & 0 \\ 
$T_3$   & 10.0612 & 1 & $T_{93,027}$  & 5.6767 & 0 \\ 
$T_4$   & 9.5787  & 1 & $T_{93,028}$  & 5.6767 & 0 \\ 
$T_5$   & 9.5257  & 1 & $T_{93,029}$  & 5.6767 & 0 \\ 
$T_6$   & 9.5025  & 1 & $T_{93,030}$  & 5.6767 & 0 \\ 
$T_7$   & 9.4217  & 1 & $T_{93,031}$  & 5.6767 & 0 \\ 
$T_8$   & 9.3727  & 1 & $T_{93,032}$  & 5.6767 & 0 \\ 
$T_9$   & 9.3537  & 1 & $T_{93,033}$  & 5.6767 & 0 \\ 
$T_{10}$  & 9.3522  & 1 & $T_{93,034}$  & 5.6767 & 0 \\ 
\bottomrule
\end{tabular}
\label{tab1} 

\end{table}

\textbf{Triple Descriptions:}

\begin{itemize}
    \item $T_1$: ('/m/06xw2', '/location/statistical\_region/gdp\_nominal\_per\_\\capita./measurement\_unit/dated\_money\_value/currency', '/m/016zxr')
    \item $T_2$: ('/m/017l4', '/olympics/olympic\_participating\_country/medals\_won./olympics/\\olympic\_medal\_honor/olympics', '/m/0gyr\_7')
    \item $T_3$: ('/m/0bsl6', '/people/person/sibling\_s./people/sibling\_relationship/sibling', '/m/07cz2')
    \item $T_4$: ('/m/03ynwqj', '/film/film/estimated\_budget./measurement\_unit/dated\_\\money\_value/currency', '/m/02h661t')
    \item $T_5$: ('/m/0fmqp6', '/film/film/film\_production\_design\_by', '/m/0ps8c')
    \item $T_6$: ('/m/0k1jg', '/education/educational\_degree/people\_with\_this\\\_degree./education/education/major\_field\_of\_study', '/m/01zcrv')
    \item $T_7$: ('/m/02vk5b6', '/award/award\_winning\_work/awards\_won./award\\/award\_honor/honored\_for', '/m/027g6p7')
    \item $T_8$: ('/m/016zxr', '/american\_football/football\_team/current\_\\roster./sports/sports\_team\_roster/position', '/m/01rs59')
    \item $T_9$: ('/m/01803s', '/sports/sports\_team/colors', '/m/024sbq')
    \item $T_{10}$: ('/m/05563d', '/base/eating/practicer\_of\_diet/diet', '/m/01n78x')
    \item $T_{93,025}$: ('/m/0jz9f', '/film/film\_distributor/films\_distributed./film/\\film\_film\_distributor\_relationship/film', '/m/02rcdc2')
    \item $T_{93,026}$: ('/m/0dq3c', '/government/government\_office\_category/officeholders./government/government\\\_position\_held/jurisdiction\_of\_office', '/m/03188')
    \item $T_{93,027}$: ('/m/01z452', '/film/film/written\_by', '/m/02m$T_4$k')
    \item $T_{93,028}$: ('/m/0845v', '/military/military\_conflict/\\combatants./military/military\_combatant\_group/combatants', '/m/0285m87')
    \item $T_{93,029}$: ('/m/0f04v', '/travel/travel\_destination/climate./travel/travel\_destination\\\_monthly\_climate/month', '/m/03\_ly')
    \item $T_{93,030}$: ('/m/095zlp', '/award/award\_winning\_work/awards\_won./\\award/award\_honor/award\_winner', '/m/06449')
    \item $T_{93,031}$: ('/m/03gwpw2', '/award/award\_ceremony/awards\_\\presented./award/award\_honor/honored\_for', '/m/084302')
    \item $T_{93,032}$: ('/m/0gqxm', '/award/award\_category/category\_of', '/m/0g\_w')
    \item $T_{93,033}$: ('/m/0pspl', '/education/educational\_institution/\\students\_graduates./education/education/major\_field\_of\_study', '/m/02h40lc')
    \item $T_{93,034}$: ('/m/09gmmt6', '/film/film/other\_crew./film/film\_crew\_gig/film\_crew\_role', '/m/0215hd')
\end{itemize}

\end{document}